\documentclass[10pt,twocolumn,letterpaper]{article}

\usepackage{cvpr} % uncomment this for the final submission
\usepackage{times}
\usepackage{epsfig}
\usepackage{graphicx}
\usepackage{amsmath}
\usepackage{lscape}
\usepackage{amssymb}
\usepackage{float}
\usepackage{tablefootnote}
\usepackage{csquotes}
\usepackage{multirow}
\usepackage{comment}
\usepackage{enumitem}
\usepackage{enumerate}
\usepackage{MnSymbol}
\usepackage{adjustbox} % Add this to your preamble
\usepackage{float}
\usepackage[table]{xcolor}

\usepackage{hyperref}
\hypersetup{
    colorlinks=true,
    linkcolor=blue,
    filecolor=magenta,      
    urlcolor=cyan,
    pdftitle={Overleaf Example},
    pdfpagemode=FullScreen,
    }
\usepackage{setspace,lipsum}
\usepackage{booktabs}

\begin{document}

%%%%%%%%% TITLE
\title{The Third Competition on Document Forgery Detection on ID-Cards and Passports}

\author{ 
Juan E. Tapia\textsuperscript{$\dagger$}\textsuperscript{1},
Mario Nieto\textsuperscript{$\dagger$}\textsuperscript{2}, 
Juan M. Espin\textsuperscript{$\dagger$}\textsuperscript{2}, 
Alvaro S. Rocamora\textsuperscript{$\dagger$}\textsuperscript{2},
Javier Barrachina \textsuperscript{$\dagger$}\textsuperscript{2},\\
Naser Damer\textsuperscript{$\dagger$}\textsuperscript{3,4}, 
Christoph Busch\textsuperscript{$\dagger$}\textsuperscript{1},
% INCODE
Aleksei Grishin\textsuperscript{*,5}, Rinat Kuzmin\textsuperscript{*,5}, Spandana Vemulapalli\textsuperscript{*,5},\\
%IDVC
Daniel Schulz\textsuperscript{*,6},  %IDVC
Sebastian Gonzalez\textsuperscript{*,6}, %IDVC
% UL-FRI
Batagelj Borut\textsuperscript{*,7},
Pahor Jure\textsuperscript{*,7},
Bri{š}ki Lea\textsuperscript{*,7}, \\
% L3i
Nicolas Sidere\textsuperscript{*,8},
Bao Quoc Dang\textsuperscript{*,8},
Jean-Christophe Burie\textsuperscript{*,8},
% Sisma
Ángela Barriga\textsuperscript{*,9}, \\
\textsuperscript{1}Hochschule Darmstadt (h-da), da/sec-Biometrics and Internet Security Research, Germany,\\
\textsuperscript{2}Facephi Biometrics, Spain,\\
\textsuperscript{3}Fraunhofer Institute for Computer Graphics Research IGD, Darmstadt, Germany,\\
\textsuperscript{4}Department of Computer Science, TU Darmstadt, Darmstadt, Germany,\\
\textsuperscript{5}Incode Technologies Inc., USA, %San Francisco, CA 94105,
\textsuperscript{6}ID VisionCenter (IDVC), Santiago, Chile,\\
\textsuperscript{7}University of Ljubljana,Ljubljana, Slovenia,
\textsuperscript{8}L3i, La Rochelle University, La Rochelle, France,\\
\textsuperscript{9}Mobbel Smart System, Spain,\\
\\
\textsuperscript{$\dagger$}{\tt\small Organizer}
\textsuperscript{$*$}{\tt\small Competitors}, %{\textbf{Full version on IEEE}}
}

\maketitle
\thispagestyle{empty}

\begin{abstract}
This paper presents a comprehensive analysis of the results from the Third International Competition on Document Forgery Detection on ID-Cards and Passports, which was held across two distinct tracks. Track 1 evaluates a synthetic-data-based ID-PAD system under controlled but diverse conditions, where the winning team, \textit{Incode}, achieves an $AV_{Rank}$ of 27.82\%, confirming consistent performance across metrics and highlighting the importance of a balanced, generalizable design. In Track 2, the challenge intensifies with heterogeneous attack scenarios across different domains, where \textit{Incode} again achieved the top position with an $AV_{Rank}$ of 68.71\% across thresholds, outperforming some baselines and established methods. These results demonstrate that PAD effectiveness requires not only high accuracy but also consistency across diverse attack types and imaging conditions. The success of this initiative across both tracks underscores the value of collaboration between companies and academic teams. This year, more than \textit{63 teams} were registered, and more than \textit{100 submission models} were evaluated. 
This competition has evolved into a leading benchmark state-of-the-art in PAD on ID documents, setting the standard for performance, reproducibility, and real-world applicability in secure identity verification.
%\vspace{-0.5cm}

\end{abstract}

\section{Introduction}
Presentation Attack Detection (PAD) is undergoing a transition driven by the convergence of physical and digital attack vectors. Easy access to multimodal generative tools (e.g., Stable Diffusion) and large language models empowers attackers to create increasingly realistic spoofs, keeping PAD in a state of continuous evolution. This shift requires PAD systems to move beyond single-modality, static defenses toward adaptive, multimodal approaches that can detect synthesized imagery, manipulated documents, passports, and coordinated social-engineering attempts. This third competition aims to provide a common platform to improve the state of the art in a challenging sequestered dataset and help to fight fraud.

The third edition marks a significant evolution in the benchmarking of identity document security. While previous editions 2024 \cite{first_challenge} and 2025 \cite{second_challenge} focused primarily on ID Cards, the 2026 challenge introduces a critical new dimension: the inclusion of Passports. This addition addresses the global need for robust authentication systems capable of handling a broader spectrum of official credentials under diverse and adversarial conditions. 

The goals of this competition are: advance the state of the art (SOTA) in the PAD-on-Document challenge by improving the generalization capabilities of algorithms and systems used by academic groups and biometric companies. Evaluate robustness against social attacks that exploit human factors and accelerate commercial adoption so citizens gain more reliable, convenient, and secure remote onboarding experiences. These conditions enable broader access to services and daily life benefits \cite{Alvaro-T-BIOM}.

The competition is structured into two distinct tracks, each designed to simulate different operational realities. %%Track 1 focuses on controlled synthetic data to lower the barrier for research and ensure privacy compliance, while Track 2 serves as an "open-set" arena, mirroring commercial deployment scenarios where developers leverage vast, heterogeneous datasets to maximize generalization.

The final performance of all systems is measured against a common, sequestered test set that remains hidden from participants. This evaluation set consists of real-world captures of bona fide, composite, print, and screen samples covering both ICAO and non-compliant document versions.

\subsection{Track 1}
Track 1 is open mainly to academic institutions and specifically designed for participants to evaluate PAD algorithms in a strictly controlled environment based on the open-set dataset. To ensure a fair comparison and adhere to data privacy standards, teams in this track are permitted to \textbf{\textit{use only the official synthetic dataset}} provided by the organizers for the training process. 

The 2026 Track 1 dataset introduces high-fidelity synthetic ID cards and Passports. In this track, bona fide samples are defined as images generated from empty templates filled with randomized user data and rendered onto a simulated PVC card or onto glossy paper. This track seeks to identify the most effective architectures and models for cross-document generalization when training data is limited to a single, high-quality synthetic source.

\subsection{Track 2}
In contrast to Track 1, Track 2 represents a more flexible and commercially realistic scenario. Here, the definition of bona fide shifts to genuine ID Cards and Passports issued by a government agency without any modification. Consequently, any reproduction or modification of the document, even on PVC or glossy paper, is classified as an attack. 

Track 2 is open to all, focusing mostly on industrial partners that have access independently to bigger datasets to represent real scenarios. Participants are encouraged to use any combination of proprietary, homemade, or open-access datasets to train their models.
 
\section{Datasets}
\label{sec:datasets}

\subsection{Track 1: Shared Synthetic Dataset}

For the first track of the competition, we provide a dataset based on synthetic document generation. The provided materials consist of two major components: a set of synthetic ID cards and a set of synthetic passports. Both subsets have been meticulously crafted to include varied ID templates, security features, and photo-realistic user portraits. 

\begin{figure*}[t]
\centering

% --- Row 1 ---
\subfloat[ESP ID Card Screen Display]{
  \includegraphics[width=40mm,height=26mm]{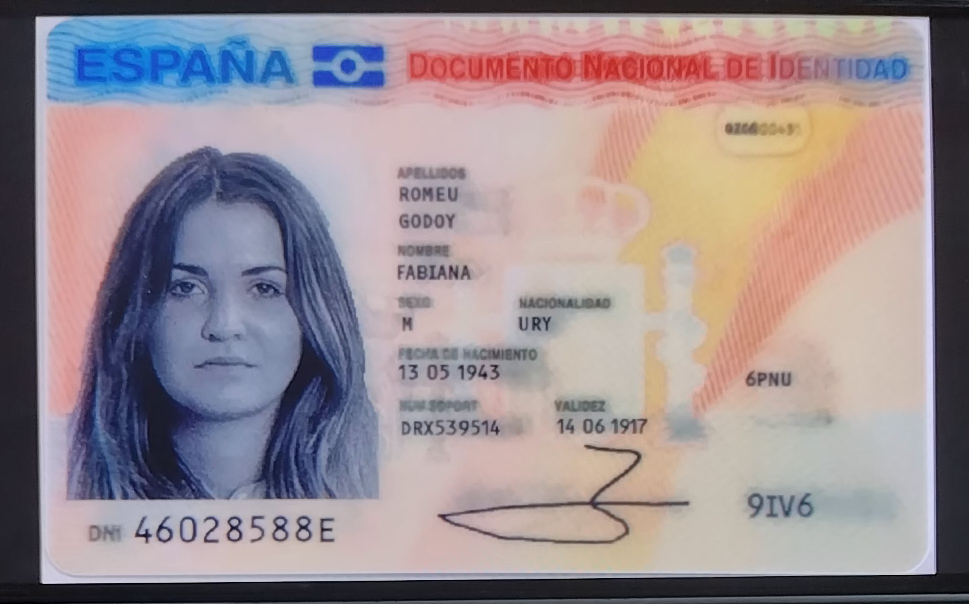}\label{fig:spoof-esp}
}
\subfloat[ESP ID Card Colour Print]{
  \includegraphics[width=40mm,height=26mm]{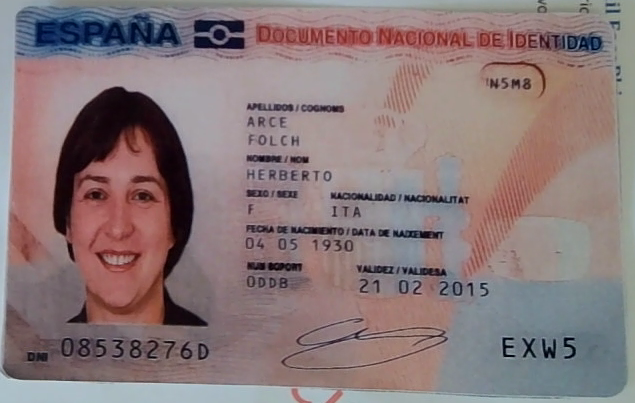}\label{fig:color-copy-esp}
}
\subfloat[ESP ID Card Physical Manipulation]{
  \includegraphics[width=40mm,height=26mm]{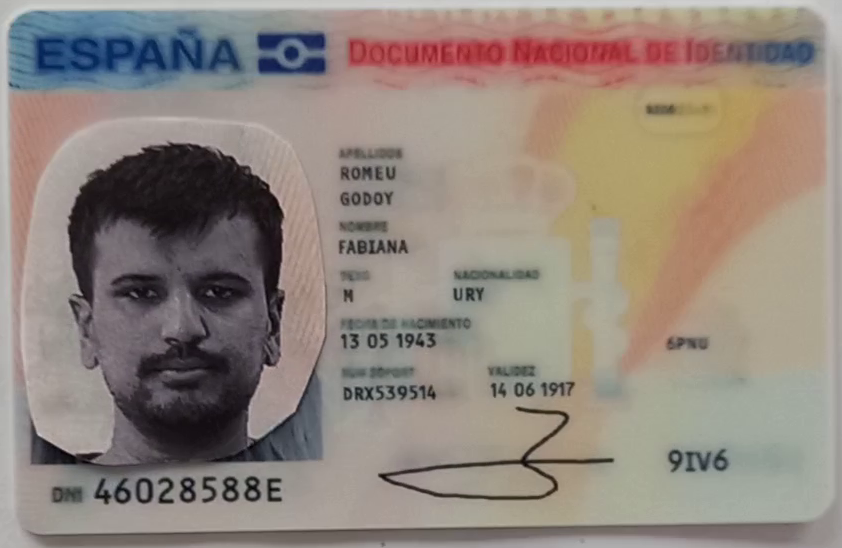}\label{fig:composite-manual-esp}
}
\subfloat[ESP ID Card Digital Manipulation]{
  \includegraphics[width=40mm,height=26mm]{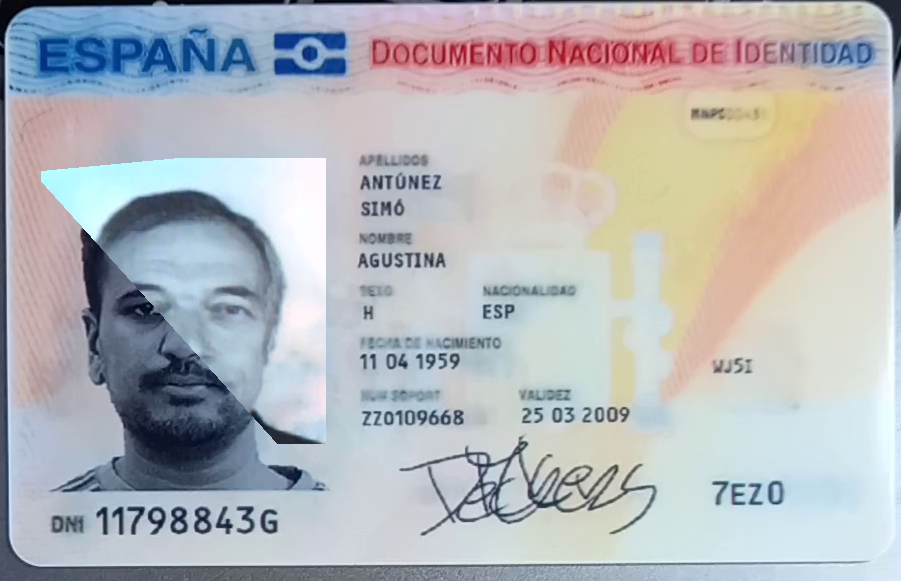}\label{fig:composite-digital-esp}}\\
% --- Row 2 ---

\subfloat[ESP ID Card Digital Bona fide]{
  \includegraphics[width=40mm,height=26mm]{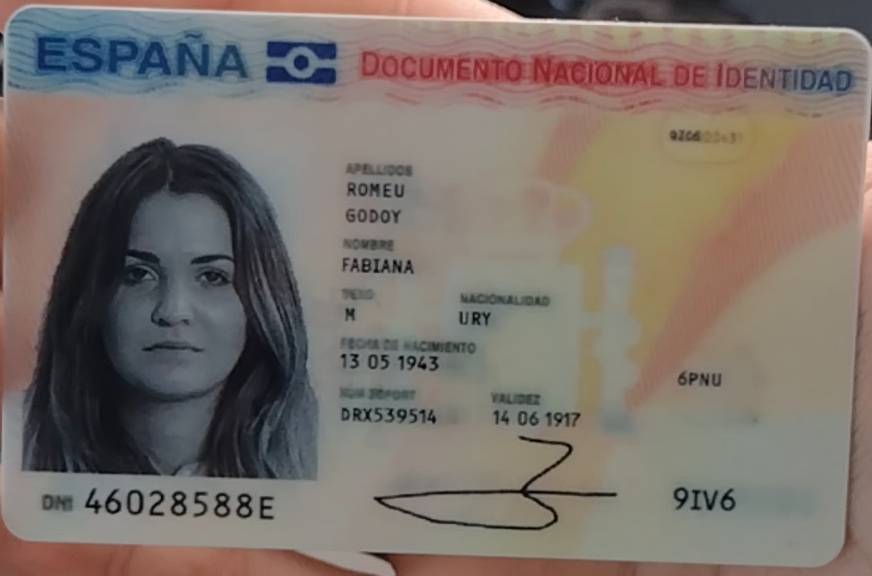}\label{fig:bona-esp}
}
\subfloat[POR Passport Synth Bona Fide]{
  \includegraphics[width=40mm,height=26mm]{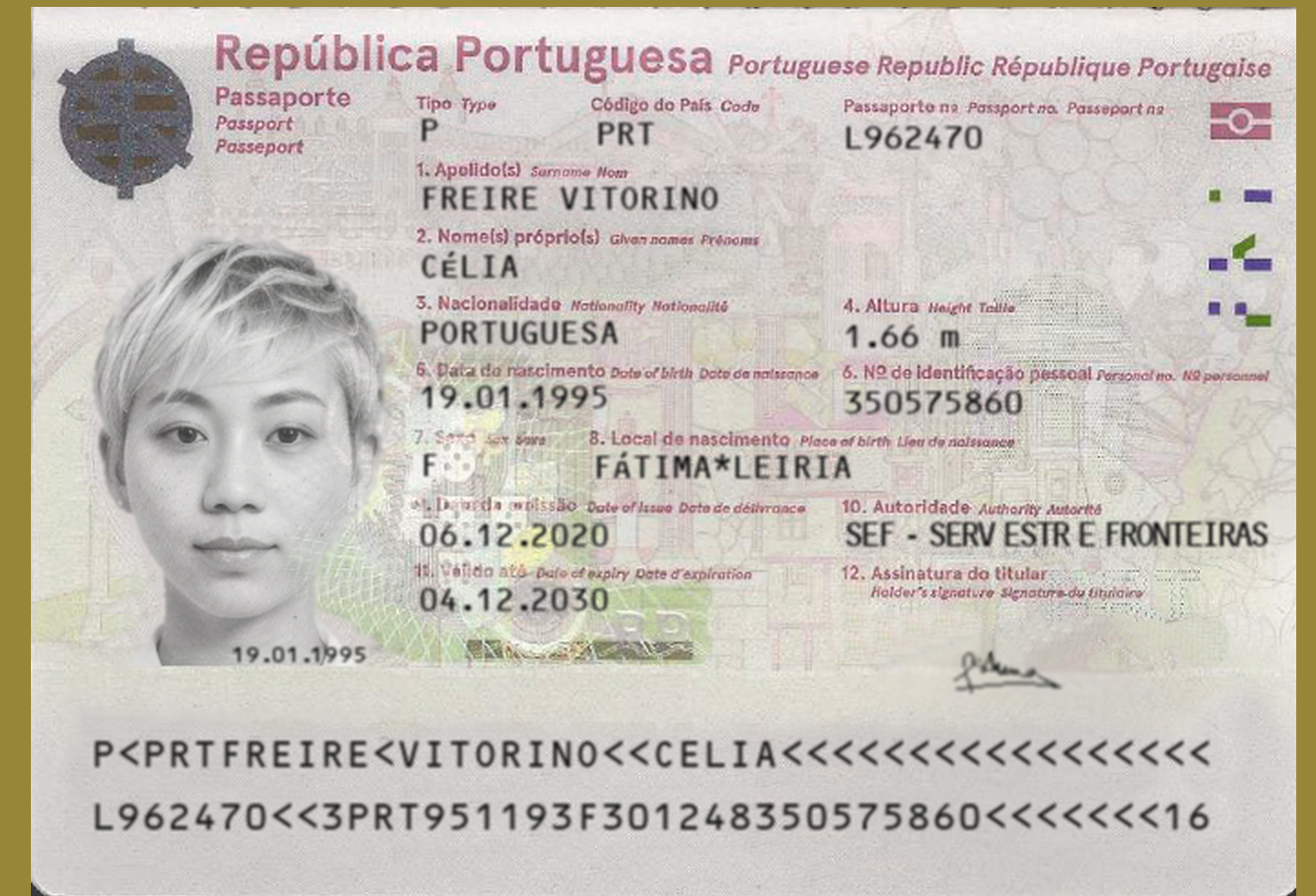}\label{fig:bonafide-por}
}
\subfloat[POR Passport Synth Screen]{
  \includegraphics[width=40mm,height=26mm]{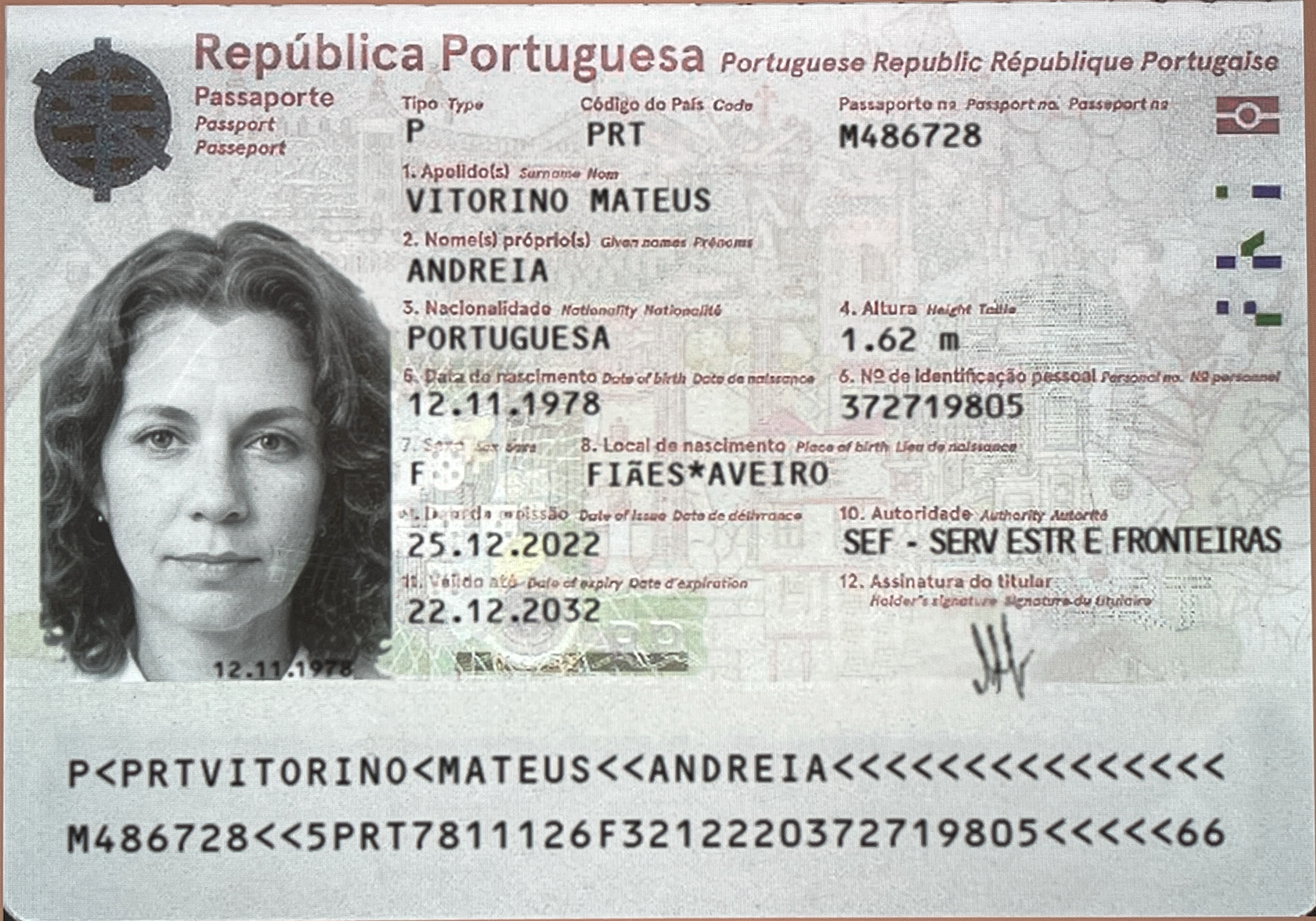}\label{fig:screen-por}
}
\subfloat[POR Passport Synth Print]{
  \includegraphics[width=40mm,height=26mm]{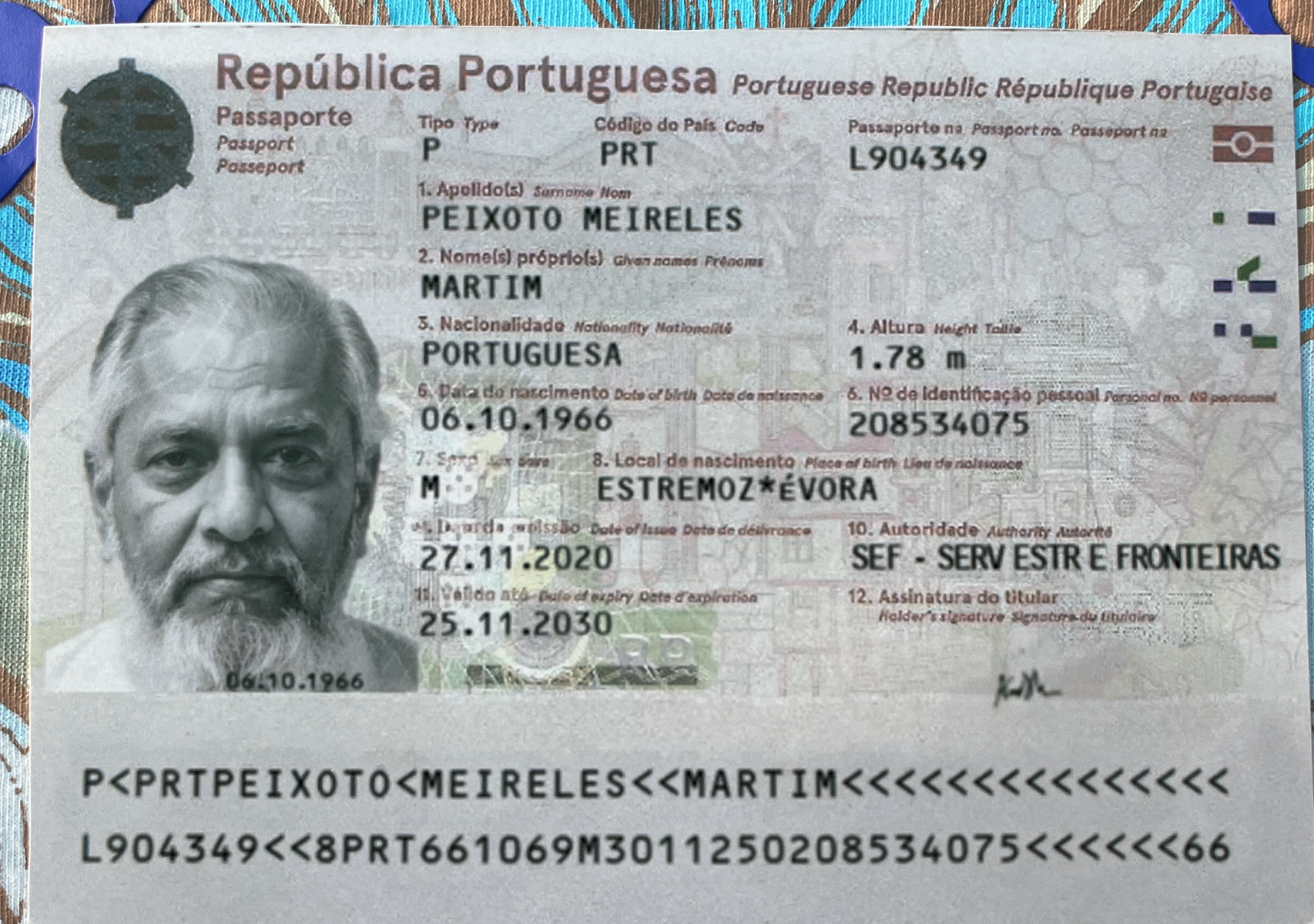}\label{fig:print-pot}
}
\begin{minipage}[b]{0.24\textwidth} % Empty placeholder to maintain alignment if needed
\end{minipage}
\begin{minipage}[b]{0.24\textwidth}
\end{minipage}

\caption{Examples of ID Cards and Passports presentation and manipulation attacks generated for the Track 1 Shared Synthetic Dataset.}
\label{fig:attacks-maded}
\end{figure*}

The full distribution of image types and associated user identities (subjects) for this dataset is detailed in Table~\ref{tab:db-track1-dist}. Also, examples of the attacks generated for the dataset can be seen in Figure \ref{fig:attacks-maded}.

% Use figure* for full-page width (top of page)

The subsequent subsections discuss the specific contents, file structures, and the rationale behind the design of each document category.

\subsection{Synthetic ID Card Dataset}

This dataset was generated using digitally generated ID card templates that mimic a wide range of identity documents from several countries. We initially collected front-facing ID templates and manually cleared all personally identifiable information, such as names, dates, faces, and signatures, using Adobe Photoshop\footnote{\url{https://www.adobe.com/es/products/photoshop.html}}. This ensured a clean and artifact-free basis for further manipulations.

Next, synthetic identities were generated using a combination of computer vision techniques and public datasets. Seamless cloning methods~\cite{poisson} were used to insert new data, including randomly selected faces, synthetic signatures, names, and alphanumeric characters, into the cleared templates. These assets were sourced from a combination of public face datasets~\cite{face-dataset-1, face-dataset-2, face-dataset-3} and signature datasets~\cite{signature-dataset-1, signature-dataset-2, signature-dataset-3}, allowing the creation of unique and realistic ID cards (see examples in Figure~\ref{fig:attacks-maded}).

The dataset provided includes 12,000 images divided between various classes representing both semi-synthetic bona fide and attack scenarios, manually generated based on the traditional process. Only the front side of the ID cards is used. The composition is as follows:

\begin{itemize}
    \item \textbf{3,000 semi-synthetic bona fide images}, created by printing ID card designs with synthetic face images on PVC cards and subsequently capturing them using mobile devices.
    \item \textbf{3,000 screen attack images}, captured by displaying the printed and recaptured PVC ID cards on a variety of digital screens (e.g., laptops, tablets, smartphones) and photographing them again in various resolutions.
    \item \textbf{3,000 print attack images}, consisting of gray-scale and color printed reproductions on glossy and standard paper with different thicknesses, and photographing them again in various resolutions.
    \item \textbf{3,000 composite attack images}, generated by physically altering the user's face in the PVC ID cards with paper cutouts of different faces, pasted in both rectangular and irregular shapes, and photographing them again in various resolutions. And, by digitally altering the synthetic ID cards, replacing facial regions with other synthetic identities using seamless technology.
\end{itemize}

Each image class in the shared dataset is uniformly distributed across a large set of semi-synthetic identities, covering \textit{24 countries} and 155 unique card templates.

This carefully annotated semi-synthetic dataset forms the foundation for participants to develop presentation attack detection models in Track 1, with a controlled and reproducible setup that reflects deployment challenges.

\subsection{SynID: Passport Synthetic Dataset}
The second component of the Track 1 dataset is the SynID Passport collection~\cite{synid_pass}. This dataset was developed using a novel hybrid methodology that integrates synthetic biometric assets with open-access document information.

The generation pipeline for the passport dataset follows a structured five-component process: template normalization from publicly available or internally generated layered Photoshop (PSD) files; generation of culturally appropriate subject metadata; selection and filtering of ICAO-compliant biometric images~\cite{caras_juan}; multimodal compositing of document layers; and reconstruction of complex visual security features such as logos and patterns. 

The complete SynID passport dataset consists of a total of $9,141$ images. This volume is derived from around $1,015$ unique synthetic individuals, for which $1,015$ high-fidelity passport images were automatically generated per country (Spain, Portugal, and Poland), resulting in $3,045$ simulated bona fide instances. To simulate realistic adversarial scenarios, the researchers manually produced presentation attacks from these synthetic base images:

\begin{itemize}
    \item \textbf{$3,045$ (1,015 per country) simulated bona fide images:} representing the original synthetic documents.
    \item \textbf{$3,049$ print attack images:} created by placing four images per sheet on glossy 180-gram paper at 600 dpi to mimic official materials, subsequently captured using two different smartphones.
    \item \textbf{$3,047$ screen attack images:} captured by displaying the generated documents on Dell laptop screens and recording them with a smartphone.
\end{itemize}

\begin{table}[h]
\caption{Distribution Track 1 Shared Synthetic Database }
\label{tab:db-track1-dist}
\resizebox{\columnwidth}{!}{%
\begin{tabular}{l|cc|cc|}
\cline{2-5}
\multirow{2}{*}{}                      & \multicolumn{2}{c|}{ID Cards}     & \multicolumn{2}{c|}{Passports}     \\ \cline{2-5} 
                                    & \multicolumn{1}{c|}{Nº Images} & Nº Unique Subjects & \multicolumn{1}{c|}{Nº Images} & Nº Unique Subjects \\ \hline
\multicolumn{1}{|l|}{Bona fide}        & \multicolumn{1}{c|}{3,000}  & 155 & \multicolumn{1}{c|}{3,045} & 1,015 \\ \hline
\multicolumn{1}{|l|}{Screen} & \multicolumn{1}{c|}{3,000}     & 155                & \multicolumn{1}{c|}{3,047}     & 1,015              \\ \hline
\multicolumn{1}{|l|}{Print} & \multicolumn{1}{c|}{3,000}     & 155                & \multicolumn{1}{c|}{3,049}     & 1,015              \\ \hline
\multicolumn{1}{|l|}{Composite} & \multicolumn{1}{c|}{3,000}  & 155 & \multicolumn{1}{c|}{-}     & -     \\ \hline
\multicolumn{1}{|l|}{\textbf{Total}}            & \multicolumn{1}{c|}{\textbf{12,000}} & \textbf{155} & \multicolumn{1}{c|}{\textbf{9,141}} & \textbf{1,015} \\ \hline
\end{tabular}%
}
\end{table}
\vspace{-0.3cm}

\subsection{Test sequestered Dataset}

The evaluation of all submitted systems, including the baseline models, was conducted using a newly curated, sequestered test set. While previous editions of the competition~\cite{first_challenge, second_challenge} utilized a static dataset to allow for direct comparisons, the third edition introduces a completely revised and expanded test set. This change is driven by the need to assess the generalization capabilities of PAD systems across varying document types, nationalities, and security features.

The current \textbf{sequestered test set} comprises approximately \textbf{27,000 images}, which constitutes one of the most challenging PAD benchmarks, for research purposes only. %That's because, depending on the track, bona fide passports are added as extras if Track 2 is being evaluated, or synthetic passports are added if Track 1 is being evaluated.

To increase the complexity and realism of the task, the dataset has been expanded beyond ID cards to include passports, significantly increasing the structural variability of the samples. Furthermore, the geographic distribution has been broadened. Instead of focusing on a fixed set of four countries, the data now encompasses a diverse range of documents from multiple nations and across various document versions. This redistribution ensures that models are evaluated on their ability to handle unseen visual and semantic patterns rather than over-fitting to specific national formats.

The dataset maintains a rigorous categorization of samples, including bona fide documents and three primary types of presentation and manipulation attacks:

\begin{itemize}
    \item \textbf{Screen attack}: Captured from a variety of digital displays (smartphones, tablets, and laptops) across multiple resolutions
    \item \textbf{Print attack}: Produced using high-quality standard paper prints and PVC card recreations.
    \item \textbf{Composite attack}: Involving physical forgeries and manual or automated digital manipulations.
\end{itemize}

To reflect the challenges of real-world acquisition, each image underwent a detailed quality capture process. The dataset incorporates significant variations in lighting conditions, image resolution, and compression artifacts. Additionally, it includes a mix of ICAO-compliant and non-compliant images. The variety is further supported by contributions from thousands of unique user IDs, ensuring a high degree of intra-class variability.

Furthermore, the dataset contains contributions from $4,000$ unique user IDs, with images spread across $25$ distinct ID versions. Notably, there is a greater presence of documents from countries such as \textit{Chile, Argentina, Costa Rica}, and \textit{Spain}. This variety ensures that models are tested against a wide range of visual and semantic variability.

Importantly, no injection attacks were considered in this evaluation. The focus remains on the physical presentation of forged identity documents, aligning with the goals of real-world deployment scenarios in online document authentication systems. Table~\ref{tab:db-track1-and-2-test} describes the distributions of the test sets for Track 1 and Track 2.

% \usepackage{graphicx}
\begin{table}[h]
\caption{Distribution for Track 1 and Track 2 Test Dataset}
\label{tab:db-track1-and-2-test}
\resizebox{\columnwidth}{!}{%
\begin{tabular}{l|cc|cc|}
\cline{2-5}
                                    & \multicolumn{2}{c|}{Track 1} & \multicolumn{2}{c|}{Track 2} \\ \cline{2-5} 
                                       & \multicolumn{1}{c|}{ID Card Imgs} & Passport Imgs & \multicolumn{1}{c|}{ID Card Imgs} & Passport Imgs \\ \hline
\multicolumn{1}{|l|}{Bona fide}     & \multicolumn{1}{c|}{2,437}      & 2,067     & \multicolumn{1}{c|}{2,437}      & 2,042     \\ \hline
\multicolumn{1}{|l|}{Screen} & \multicolumn{1}{c|}{2,083}      & 2,207     & \multicolumn{1}{c|}{2,083}      & 2,029     \\ \hline
\multicolumn{1}{|l|}{Print} & \multicolumn{1}{c|}{5,928}      & 2,133     & \multicolumn{1}{c|}{5,928}      & 2,107     \\ \hline
\multicolumn{1}{|l|}{Composite} & \multicolumn{1}{c|}{7,237}           & 2,056            & \multicolumn{1}{c|}{7,237}           & 3,994            \\ \hline
\multicolumn{1}{|l|}{\textbf{Total}}         & \multicolumn{1}{c|}{\textbf{17,685}}     & \textbf{8,463}     & \multicolumn{1}{c|}{\textbf{17,685}}     & \textbf{10,172}    \\ \hline
\end{tabular}%
}
\end{table}

Finally, it is important to note that, when evaluating Track 1, synthetic samples of PVC-printed ID cards and paper-printed synthetic passports are treated as bona fide samples. However, when evaluating Track 2, these samples are classified as attacks and are considered spoofed.

\section{Submission Process}
\label{sec:submission_process}
The competition utilized a customized platform that managed participant registration and submission logistics, forwarding evaluation queries to a dedicated private server. Such a setup facilitated the processing of multiple iterations per participant; specifically, we processed more than $100$ submissions. Only 20 teams of the 63 registered submitted a valid model. The organizer assigned two partitions of an \textit{NVIDIA A100 GPU} with $20GB$ of VRAM each to evaluate submissions.

An API embedded in a Docker image was used to evaluate submissions. This interface features a specific endpoint that accepts an image and returns a continuous liveness score in the $[0, 1]$ interval, where $0$ indicates a presentation attack and $1$ denotes a high-confidence bona fide document. Our evaluation pipeline assumes all dataset images are processable; consequently, any execution errors are assigned a score of $0$, effectively detected as attacks.

%After the API processes the entire dataset, the performance metrics defined in Section~\ref{sec:eval} are computed. While the leaderboard is updated to reflect only the highest-ranking submission per participant, these peak results are the ones documented in this paper.

\section{Performance Evaluation Criteria}
\label{sec:eval}

The evaluation of biometric PAD systems in this competition follows the international standards defined in ISO/IEC 30107-3\footnote{\url{https://www.iso.org/standard/79520.html}}. To quantify classification performance, we focus on the Attack Presentation Classification Error Rate (APCER), the Bona fide Presentation Classification Error Rate (BPCER), and specific operational points denoted as BPCER\textsubscript{AP}. These metrics assess the system's ability to differentiate between bona fide samples and various Presentation Attack Instrument Species (PAIS).

The APCER represents the rate at which attack attempts are erroneously accepted as bona fide presentations. It is calculated independently for each PAIS as defined in Equation~\ref{eq:apcer}, where $N_{PAIS}$ denotes the total number of attack samples for a specific category. The variable $RES_{i}$ is a binary indicator: it is $1$ if the $i$th sample is correctly identified as an attack and $0$ if it is misclassified as bona fide based on a set decision threshold.

\begin{equation}\label{eq:apcer}
    {APCER_{PAIS}}=1 - \frac{1}{N_{PAIS}}\sum_{i=1}^{N_{PAIS}}RES_{i}
\end{equation}

On the other hand, the BPCER quantifies the frequency with which legitimate users are falsely rejected by the system. As shown in Equation~\ref{eq:bpcer}, this metric is computed using the total number of bona fide samples ($N_{BF}$), where $RES_{i}$ follows the same binary logic described above. Together, APCER and BPCER characterize the diagnostic capability of the system at a given threshold. 

\begin{equation}\label{eq:bpcer}
    BPCER=\frac{\sum_{i=1}^{N_{BF}}RES_{i}}{N_{BF}}
\end{equation} 

To provide a comprehensive view of system behavior across different operational requirements, we utilize the Equal Error Rate (EER) and BPCER\textsubscript{AP}. The EER identifies the threshold where APCER and BPCER are equivalent, visually represented by the intersection of the Detection Error Trade-off (DET) curve with the diagonal identity line. Furthermore, the BPCER\textsubscript{AP} metric indicates the BPCER value when the APCER is constrained to a fixed target of $100/AP$. In this study, we report BPCER\textsubscript{10}, BPCER\textsubscript{20}, and BPCER\textsubscript{100}, which correspond to APCER operating points of 10\%, 5\%, and 1\%, respectively.

The final standing of the participants is determined by an average weighted ranking, $AV_{Rank}$. This cumulative metric assigns higher significance to more stringent operational points to reflect the demands of high-security applications. Specifically, we apply weighting factors of $0.2$ for BPCER\textsubscript{10}, $0.3$ for BPCER\textsubscript{20}, and $0.5$ for BPCER\textsubscript{100}. The winning participant is the one achieving the lowest $AV_{Rank}$ value, computed as follows:
\vspace{-0.3cm}

\begin{equation}\label{eq:avrank}
\scriptstyle
    AV_{Rank}=BPCER_{10}\times0.2+ BPCER_{20}\times0.3 + BPCER_{100}\times0.5
\end{equation} 

\section{Baseline Description}
\label{sec:baseline}
This section details the architectural designs and strategies employed for both the baseline models and the participant entries across Tracks 1 and 2. According to the state-of-the-art~\cite{GONZALEZ-PR, markham2023openset, first_challenge, second_challenge, Alvaro-T-BIOM}.

\subsection{Baselines for Track 1}
Based on the results of the previous year \cite{second_challenge}, the baseline was updated. The convolutional approach from the previous edition has been replaced by a modern vision transformer-based architecture (ViT) for the challenge.

The baseline established for the 2025 edition focused on a controlled training scenario using the official synthetic ID card dataset. The architecture selected was \textit{EfficientNetV2-S}~\cite{efficientnetv2}. At the time, a convolutional neural network (CNN) was preferred over ViT due to the limited amount of available training data; CNNs like \textit{EfficientNet} tend to exhibit greater efficiency in low-data regimes.

This year, for the 2026 competition, the baseline has been updated to reflect the trend toward more robust feature extractors, in line with the expanded dataset and last year’s winning proposal \cite{second_challenge}.

Although the preprocessing steps, the 5\% padding, and the data augmentation strategies remain identical to those in the 2025 configuration to ensure procedural consistency, two significant updates have been implemented in addition to adjusting the input size to the network to $224 \times 224$:

\begin{itemize}
\item \textbf{Architectural Shift:} The previous CNN architecture has been replaced with \textit{DINOv2} (Variant B)~\cite{dinov2}. Leveraging the powerful self-supervised pre-training of the \textit{DINOv2} framework. This ViT backbone provides superior spatial representations that are particularly effective for detecting subtle textural artifacts in document forgeries.
\item \textbf{Inclusion of Synthetic Passports:} To address the goal of cross-document generalization, the training and validation sets were expanded. In addition to the synthetic ID card data used in 2025, the \textit{SynID: Passport Synthetic Dataset} was integrated into the training pipeline. This expansion significantly increases the structural variability of the ''bona fide'' and ''attack'' classes, forcing the model to learn features that are invariant to document type (ID card vs. passport). The training and validation sets were constructed by partitioning the data based on unique document identities per class, using a 40/60\% split
\end{itemize}

The training follows the same four-class classification objective, but the inclusion of passport samples alongside ID cards allows this 2026 baseline to serve as a more rigorous benchmark for generalization across document categories and national versions. Table~\ref{tab:track1_baseline_2026} provides a summary of the ID Card and Passport dataset distribution used for the training process of Track 1 Baseline of 2026 Challenge:

% \usepackage{graphicx}
\begin{table}[H]
\caption{Baseline 2026 Track 1 - Train and Validation Distribution}
\label{tab:track1_baseline_2026}
\resizebox{\columnwidth}{!}{%
\begin{tabular}{l|cc|cc|}
\cline{2-5}
                                       & \multicolumn{2}{c|}{Train}         & \multicolumn{2}{c|}{Validation}    \\ \cline{2-5} 
 & \multicolumn{1}{c|}{ID Card Imgs} & Passport Imgs & \multicolumn{1}{c|}{ID Card Imgs} & Passport Imgs \\ \hline
\multicolumn{1}{|l|}{Bona fide}        & \multicolumn{1}{c|}{2,397} & 1,218 & \multicolumn{1}{c|}{603}   & 1,827 \\ \hline
\multicolumn{1}{|l|}{Screen}    & \multicolumn{1}{c|}{2,351} & 1,219 & \multicolumn{1}{c|}{649}   & 1,828 \\ \hline
\multicolumn{1}{|l|}{Print}     & \multicolumn{1}{c|}{2,407} & 1,220 & \multicolumn{1}{c|}{593}   & 1,829 \\ \hline
\multicolumn{1}{|l|}{Composite} & \multicolumn{1}{c|}{2,451} & -     & \multicolumn{1}{c|}{549}   & -     \\ \hline
\multicolumn{1}{|l|}{\textbf{Total}}            & \multicolumn{1}{c|}{\textbf{9,606}} & \textbf{3,657} & \multicolumn{1}{c|}{\textbf{2,394}} & \textbf{5,484} \\ \hline
\end{tabular}%
}
\end{table}

\subsection{Baselines for Track 2}

For the 2026 edition of the competition, the Track 2 baseline framework has been completely changed. Moving away from the unified single-network approach utilized in the 2025 edition, this year's primary baseline transitions to a dual-network ensemble system. This architectural pivot is directly motivated by the winning strategy of the previous competition edition~\cite{second_challenge}, and for the work published by Gonzalez et al.~\cite{GONZALEZ-PR}. Both demonstrated that decoupling different attack modalities significantly improves overall attack detection.

While the underlying training dataset distribution remains strictly identical to the 2025 baseline configuration, summarized in Table~\ref{tab:db-baseline-exp1}, the 2026 system discards the traditional single-stage multi-class pipeline. Instead, it deploys two specialized networks running in parallel, each engineered to intercept distinct physical or semantic architectural vulnerabilities:
\begin{itemize}
    \item \textbf{Presentation Attack Specialist:} The first independent network is dedicated entirely to physical recapture traces. It focuses strictly on detecting print and screen attacks, learning to localize high-frequency artifacts.
    \item \textbf{Manipulation and Forgery Specialist:} The second independent network isolates composite attacks instead of looking for physical printing artifacts; it targets semantic and structural inconsistencies within the document, such as physical cut-and-paste lines or digital face-and-text swapping operations.
\end{itemize}

% \usepackage{graphicx}
\begin{table}[h]
\scriptsize
\centering
\caption{Baseline 2026 Track 2 - Train and Validation distribution from a private dataset}
\label{tab:db-baseline-exp1}

\begin{tabular}{l|c|c|c|}
\cline{2-4}
                                       & Train   & Val    & Total   \\ \hline
\multicolumn{1}{|l|}{Bona fide}        & 64,933  & 11,458 & 76,391  \\ \hline
\multicolumn{1}{|l|}{Screen}    & 82,232  & 14,511 & 96,743  \\ \hline
\multicolumn{1}{|l|}{Print}     & 30,398  & 5,364  & 35,762  \\ \hline
\multicolumn{1}{|l|}{Composite} & 90,101  & 15,900 & 106,001 \\ \hline
\multicolumn{1}{|l|}{\textbf{Total}}            & \textbf{267,664} & \textbf{47,233} & \textbf{314,897} \\ \hline
\end{tabular}%

\end{table}

Both stages employ the \textit{MobileNetV2} architecture~\cite{sandler2018mobilenetv2}. Rather than fine-tuning pre-existing weights, these networks are trained entirely from scratch on our data.

To support this specialized dual-network strategy, the preprocessing pipeline scales up technical requirements from the previous year. The input dimension is increased to a higher resolution of $448\times448$. 

Before passing to the ensemble, a semantic segmentation process isolates the identity document, removing background pixels to guarantee that both models optimize exclusively on document features. An aggressive data augmentation strategy is continuously applied during training via the \textit{imgaug} library. This includes color space transformations, affine and perspective warping to replicate arbitrary capture angles, and sensory noise distortions.

Finally, in the final execution stage, the continuous score outputs of these specialized classifiers are concatenated to yield a singular, highly robust unified PAD decision. By pivoting the 2026 baseline around an ensemble of specialized deep networks trained across globally diverse private data, this edition establishes a notably more rigorous and resilient benchmark against simultaneous manipulation and presentation threats.

\section{Submission to the Competition}
For this competition, 63 teams were registered. The systems of three teams with the lower $AV_{Rank}$ are described for each track. All the submission results are available in the \textit{\textbf{supplementary material}} associated with this work and on the official website \footnote{\url{https://tinyurl.com/38pxppw9}}.

\subsection{Track 1 and 2 - First Position - Incode}
\textbf{Track 1}: The team Incode presents IDV\_103, a PAD system trained on the dataset provided by the organizers for Track 1. The approach employs a two-stage preprocessing pipeline: first, \textit{YOLOv11n}~\cite{khanam2024yolov11} is used to detect the minimal bounding box containing the identity document in the input photo; then both cropped document regions and full images are used as model inputs to preserve local document details as well as global visual context.

The core model is based on \textit{ConvNeXt-Base}~\cite{convNext} initialized with \textit{DINOv3}~\cite{simeoni2025dinov3} weights and trained as a multilabel binary classifier, predicting the presence of each PA type independently. To improve robustness and increase attack diversity during training, the system uses both cropped and full-image augmentations. In addition, synthetic presentation attack samples (SPAS)~\cite{spas} are generated from the bona fide images using the \textit{FLUX Klein 9B model}~\cite{flux-2-2025}, enriching the training data with more diverse spoofing patterns. Training with SPAS takes $2.83$ points of the on the evaluation dataset, compared to the $IDV\_101$ version, which is exactly the same setup without SPAS.

For post-processing, the final prediction is produced using test-time augmentation and an ensemble of models trained on different folds, which improves stability and generalization across unseen samples. This combination of document localization, multilabel training approach, synthetic attack generation, TTA, and fold-based ensembling enables strong generalization across diverse presentation attack types.

\textbf{Track 2}: The team \textit{Incode} presents \textit{IDV\_301}, a presentation attack detection system trained and evaluated on \textit{Incode's} private in-house dataset ($100K$ bona fide document samples and $70K$ attack samples) for the second track. The approach employs a two-stage preprocessing pipeline: first, an in-house document detection model is used to detect and crop the identity document region in the input photo; then, both cropped document regions and full images are used as model inputs to preserve local document details as well as global visual context.

The core model is based on \textit{ConvNeXt-Tiny}~\cite{convNext} initialized with ImageNet-pretrained weights and trained as a multilabel binary classifier, predicting the presence of each presentation attack type independently. To improve robustness during training, the system uses both cropped and full-image augmentations, allowing the model to learn complementary cues from localized document regions and the surrounding acquisition context.

Unlike the Track 1 system, this version does not rely on synthetic PA data or fold-based ensembling. The final prediction is produced by a single \textit{ConvNeXt-Tiny} model with test-time augmentation applied at inference time. This design provides an efficient and robust solution while maintaining strong generalization across diverse presentation attack types. The combination of in-house document localization, multilabel training, and TTA enables reliable presentation attack detection on the private evaluation dataset.

\subsection{Track 1 - Second Position - UL-FRI}
The team UL-FRI presents a multi-class presentation attack detection model based on a \textit{LoRA} fine-tuned \textit{DINOv3} ViT-L backbone. The model was trained on the given Track 1 datasets. The preprocessing pipeline consists of SAM3-based card detection using candidate scoring based on aspect ratio and \textit{InsightFace} buffalo\_l face detection, with detection fallback based on \textit{YOLO11n-based} object detection. The cropped cards are subsequently rotated to a landscape orientation, normalized, and zero-padded to the target input resolution of $512\times512$ pixels. Augmentation includes random contrast, brightness, hue variation, random geometric transforms, and random \textit{JPEG} compression augmentation of varying quality. The vision foundation model representing the backbone of the model is fine-tuned using \textit{LoRA} with a low rank of $r=2$ and dropout regularization to mitigate overfitting. The classification head consists of a two-layer fully-connected network utilizing batch normalization, \textit{GELU} activation, and \textit{dropout}. The model was trained using Adam optimizer with weight decay and $5e-5$ learning rate for $4$ epochs with weighted cross-entropy. 

%\subsection{Track 1 - Third Position - L3i}
\subsection{Track 1 - Third Position - L3i}
The \textbf{L3i} team presents \textbf{SwinDCT-PAD} (Swinc1), a dual-stream model for identity document presentation attack detection on the IJCB 2026 Track 1 dataset ($6,045$ bona fide, $20,520$ attacks). To ensure cross-domain robustness, a hybrid data split is used: passport samples are partitioned by country of origin (preventing domain leakage), while digital composite attacks are split at the instance level. Training employs random resized crops, color jittering, Gaussian blur, and $40\%$ coarse dropout to mitigate layout biases.

The architecture combines a spatial stream based on a \textit{LoRA}-enhanced \textit{Swin-TransformerV2} and a spectral stream using the \textit{Discrete Cosine Transform} (DCT). Optimized via \textit{AdamW} ($\eta = 5 \times 10^{-5}$, weight decay $= 0.05$) for 30 epochs, the model minimizes a multi-task loss: $\mathcal{L}_{\text{total}} = \mathcal{L}_{\text{WCE}} + \gamma \mathcal{L}_{\text{BCE}}$, where class-weighted cross-entropy (with label smoothing $= 0.1$) addresses class imbalance, and an auxiliary BCE loss ($\gamma = 0.3$) regularizes a binary image quality head.

A \textit{YOLO} detector ($\text{conf} \ge 0.85$) crops the document region during preprocessing and inference. The region is padded by $5\%$, rotated if portrait, and resized to $384 \times 384 \times 3$. Inference uses a relative margin score:
\[
S = \text{clamp}\left(\frac{1}{2}\left(P_{\text{bona fide}} - \max_{i \in \mathcal{A}} P_{\text{attack}, i} + 1.0\right),\ 0.0,\ 1.0\right),
\]
where $\mathcal{A} = \{\text{print}, \text{screen}, \text{composite}\}$. This enforces robust decision boundaries, enabling strong generalization across physical and digital spoofing modalities.

%The team \textbf{L3i} presents \textbf{SwinDCT-PAD} (Swinc1) for identity document presentation attack detection (PAD) on the IJCB 2026 Track-1 dataset comprising $6,045$ bona fide and $20,520$ attack samples. To evaluate cross-domain generalization accurately, a robust hybrid data-splitting strategy is executed: passport samples from the bona fide, print, and screen replay classes are partitioned strictly by their country of origin to prevent domain leakage. While digital composite attacks are split at the instance level. The training phase is further regularized using random resized crops, color jittering, Gaussian blur, and a $40\%$ coarse dropout to reduce country-specific layout biases. 
Architecturally, \textit{SwinDCT-PAD} features a parallel dual-stream design: a spatial stream powered by a \textit{LoRA}-integrated \textit{Swin-TransformerV2}, and a spectral stream driven by a \textit{Discrete Cosine Transform} (DCT) branch. The network is optimized via \textit{AdamW} ($\eta = 5 \times 10^{-5}$, weight decay $= 0.05$) for $30$ epochs, minimizing a multi-task objective: $\mathcal{L}_{\text{total}} = \mathcal{L}_{\text{WCE}} + \gamma \mathcal{L}_{\text{BCE}}$, fusing class-weighted cross-entropy with an auxiliary regression loss. 

%The optimization objective employs a compound multi-task loss function, {integrating a 4-way class-weighted Cross-Entropy loss} with label smoothing ($0.1$) {to explicitly address class imbalance across the four distinct categories}---applying higher penalties to elusive spoofing classes---and an auxiliary Binary Cross-Entropy (BCE) loss tasked with regularizing a dynamic, binary image quality classification head via an adjustable task weight parameter ($\gamma=0.3$). Additionally, a \textit{YOLO} object detector ($\text{conf} \ge 0.85$) is deployed consistently across both offline dataset preprocessing and online real-time inference pipelines to dynamically crop the primary document area. The localized region is expanded with a $5\%$ padding margin to preserve edge artifacts, counter-clockwise rotated by $90^\circ$ if portrait-oriented, and resized to a fixed tensor shape of $384 \times 384 \times 3$. During inference, a continuous relative margin score $S$ is calculated from output logits using a localized Softmax post-processing rule: $S = \text{clamp}(\frac{1}{2}(P_{\text{bona fide}} - \max_{i \in \mathcal{A}} P_{\text{attack}, i} + 1.0), 0.0, 1.0)$, where $\mathcal{A} = \{\text{print}, \text{screen}, \text{composite}\}$. This formulation enforces robust decision boundaries, delivering high generalization against diverse physical and digital document spoofing modalities.

\subsection{Track 2 - Second Position - IDVC}

For the second track, the \textit{idcard\_pad\_st method} consists of two stages. The first stage is a document-agnostic ID card and passport detector based on \textit{YOLOv3}~\cite{yolov3}, trained on the \textit{MIDV-2020}~\cite{MIDV2020AC} dataset and a private dataset containing Chilean ID cards. The second stage performs PAD using the cropped documents from the first stage as input. This method employs a \textit{Swin-TransformerV2}~\cite{swimtransformers}, a hierarchical vision transformer that processes images efficiently by using window-based self-attention mechanisms and shifted windows to improve scalability and performance.
The model was trained on a private dataset of $152,664$ images, comprising $30K$ bona fide and $122K$ attack samples. The attacks include manual and automatic composites, prints, and screens. The majority of the data consists of Chilean ID cards, with a significantly smaller subset (two orders of magnitude smaller) of ID cards and passports from six other countries. To overcome the lack of samples for these underrepresented countries, we applied an aggressive oversampling strategy. The input for the \textit{Swin-TransformerV2} is a $224\times224$ image, which undergoes various data augmentations during training, such as 90-degree rotations, brightness and contrast adjustments, and translations.

\subsection{Track 2 - Third Position - Sisma}
For Track 2, the team \textit{Sisma} develops a PAD model for ID card verification, using a combined transformer-based approach to strengthen feature extraction and improve generalization. The approach uses ID card detection and cropping followed by resizing to a fixed input resolution, augmented with horizontal flipping, brightness and contrast adjustments, \textit{JPEG} compression artifacts, Gaussian filtering, and hue shifts, to improve robustness under diverse capture conditions. The model is trained in a supervised setup with class balancing and a weighted cross-entropy loss over bona fide, screen attack, print attack, and composite attack classes. A threshold-based decision rule is then applied at inference time to derive the final PAD score and improve operational stability across varied attack types.

\begin{table*}[ht]
\centering
\footnotesize % Reduced from \scriptsize
\caption{\label{track1-track2}Track 1 and Track 2 summary challenge results. All results are in \%.}

\begin{minipage}[t]{0.48\linewidth}
\centering
\caption*{Track 1 summary challenge results}
\begin{adjustbox}{max width=\linewidth, center}
\begin{tabular}{@{}lllllll@{}}
\toprule
Rank & Team                          & EER   & BPCER10 & BPCER20 & BPCER100 & AVRank \\ \midrule
1 &
  \cellcolor[HTML]{EFEFEF}\textbf{Incode} &
  \cellcolor[HTML]{EFEFEF}\textbf{8.42} &
  \cellcolor[HTML]{EFEFEF}\textbf{6.01} &
  \cellcolor[HTML]{EFEFEF}\textbf{15.38} &
  \cellcolor[HTML]{EFEFEF}\textbf{44.01} &
  \cellcolor[HTML]{EFEFEF}\textbf{27.82} \\
-    & DLVCLab                        & 10.68 & 12.21   & 26.78   & 55.79    & 38.37  \\
2    & UL-FRI                        & 14.27 & 21.92   & 36.01   & 57.99    & 44.18  \\
3    & L3i                           & 19.85 & 34.39   & 46.63   & 64.04    & 52.89  \\
4    & UNLJ-FRI-FE                   & 17.65 & 34.55   & 51.86   & 64.00    & 54.47  \\
5    & ArogyaPandit & 19.83 & 36.29   & 49.33   & 68.52    & 56.32  \\
6    & \cellcolor[HTML]{f8f8f8}Baseline-2026                 & 20.54 & 37.24   & 50.05   & 70.09    & 57.51  \\
7    & IDVC-PAD-IDCARD               & 20.85 & 38.85   & 55.43   & 76.51    & 62.65  \\
8    & ACROTEAM                      & 28.05 & 54.20   & 63.97   & 72.41    & 66.24  \\
9    & QualPad                       & 23.86 & 41.31   & 60.01   & 87.99    & 70.26  \\
10   & BioPAD                        & 26.71 & 45.92   & 64.36   & 91.27    & 74.13  \\
11   & SpoofSense.ai                 & 14.27 & 42.36   & 73.49   & 95.08    & 78.06  \\
12   & Become\_Digital               & 24.41 & 58.93   & 73.95   & 90.27    & 79.11  \\
13   & \cellcolor[HTML]{f8f8f8}Baseline-2025                 & 29.91 & 59.80   & 72.59   & 91.34    & 79.40  \\
14   & DLmath                        & 31.24 & 62.68   & 74.18   & 90.43    & 80.00  \\
15   & ShareID.ai                    & 46.27 & 68.28   & 77.59   & 91.12    & 82.49  \\
16   & HV                            & 31.40 & 63.30   & 81.69   & 96.55    & 85.44  \\
17   & Trufid                        & 33.16 & 69.75   & 82.33   & 96.07    & 86.69  \\
18   & pahadis                       & 41.73 & 80.01   & 88.10   & 96.73    & 90.80  \\
-   & Dermalog                       & 50.00 & 100.00   & 100.00   & 100.00    & 100.00  \\ \bottomrule
\end{tabular}
\end{adjustbox}
\end{minipage}
\hfill
\begin{minipage}[t]{0.48\linewidth}
\centering
\caption*{Track 2 summary challenge results}
\begin{adjustbox}{max width=\linewidth, center}
\begin{tabular}{@{}lllllll@{}}
\toprule
Rank & Team & EER   & BPCER10 & BPCER20 & BPCER100 & AVRank \\ \midrule
1 &
  \cellcolor[HTML]{EFEFEF}\textbf{Incode} &
  \cellcolor[HTML]{EFEFEF}\textbf{26.52} &
  \cellcolor[HTML]{EFEFEF}\textbf{56.53} &
  \cellcolor[HTML]{EFEFEF}\textbf{66.24} &
  \cellcolor[HTML]{EFEFEF}\textbf{75.06} &
  \cellcolor[HTML]{EFEFEF}\textbf{68.71} \\
2    & \cellcolor[HTML]{f8f8f8}Baseline-2026                & 26.38 & 52.65   & 66.47   & 81.13    & 71.04  \\
-    & DLVCLab*               & 26.70 & 43.72   & 58.18   & 100.00    & 76.20  \\
3    & IDVC-PAD-IDCARD               & 27.90 & 63.07   & 79.04   & 94.62    & 83.63  \\
4    & \cellcolor[HTML]{f8f8f8}Baseline-1-2025               & 33.71 & 70.08   & 82.32   & 91.92    & 84.67  \\
5    & Sisma                         & 28.27 & 68.03   & 81.92   & 94.89    & 85.62  \\
6    & UNLJ-FRI-FE                   & 29.58 & 70.98   & 82.52   & 94.35    & 86.13  \\
7    & UL-FRI                        & 31.21 & 66.26   & 80.64   & 97.41    & 86.15  \\
8    & QualPad                       & 39.94 & 72.16   & 82.16   & 94.42    & 86.29  \\
9    & HV                            & 31.21 & 71.42   & 83.88   & 96.58    & 87.74  \\
10   & BioPAD                        & 35.53 & 74.41   & 85.87   & 96.38    & 88.83  \\
11   & ArogyaPandit & 37.75 & 76.36   & 86.38   & 97.10    & 89.73  \\
12   & L3i                           & 36.41 & 77.70   & 88.23   & 97.30    & 90.66  \\
-   & UversumAI*                           & 50.07 & 93.21   & 96.65   & 99.49    & 97.38  \\\bottomrule
\end{tabular}
\end{adjustbox}
\end{minipage}
\end{table*}

\section{Results}

\textbf{Competition Winners:} The third edition of the competition on Document forgery detection on ID-Card and Passport has one winner; the same team has won both tracks. For Track 1, the \enquote{Incode} team wins, with an $AV_{Rank}$ of 27.82\% using a \textit{ConvNeXt-Base} trained as a multi-label binary classifier. For Track 2, the \enquote{Incode} team wins with an $AV_{Rank}$ of 68.71\% using a similar model to Track 1, but a Tiny version, trained on 170k samples from a private dataset. All the results are summarized in Table \ref{track1-track2}.
%%%%%AQUI track 1

%%%

% Use figure* for full-page width (top of page)
\begin{figure*}[t]
\centering

% --- Row 1 ---
\subfloat[\textbf{Global} DET - ID Card + Passport]{
  \includegraphics[width=40mm,height=40mm]{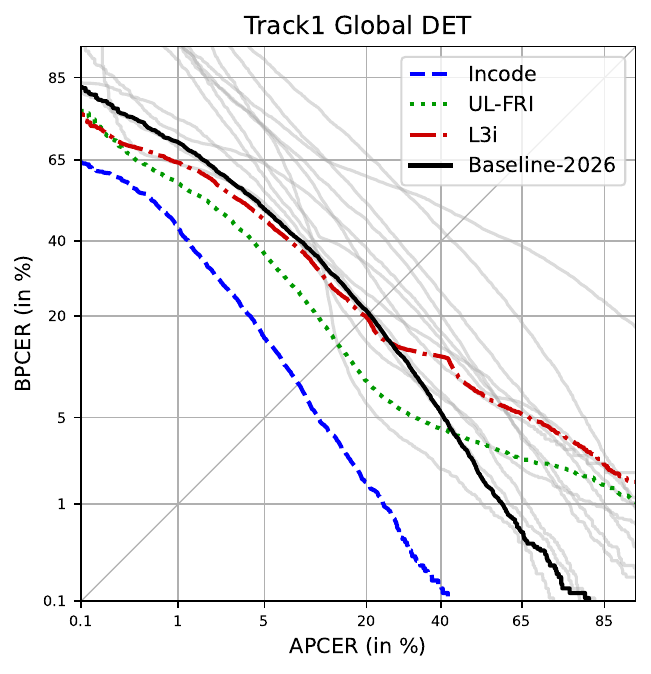}\label{fig:track1-global-det}
}
\subfloat[\textbf{Screen} DET - ID Card + Passport]{
  \includegraphics[width=40mm,height=40mm]{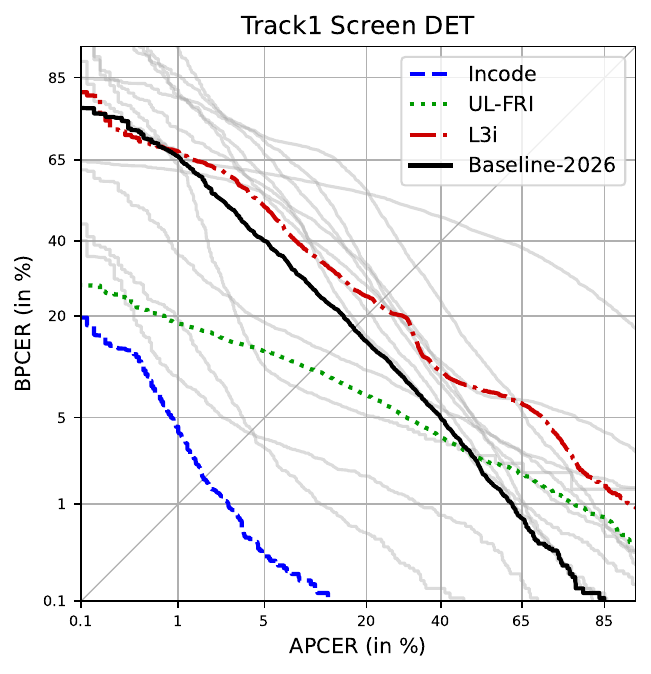}\label{fig:track1-screen-det}
}
\subfloat[\textbf{Print} DET - ID Card + Passport]{
  \includegraphics[width=40mm,height=40mm]{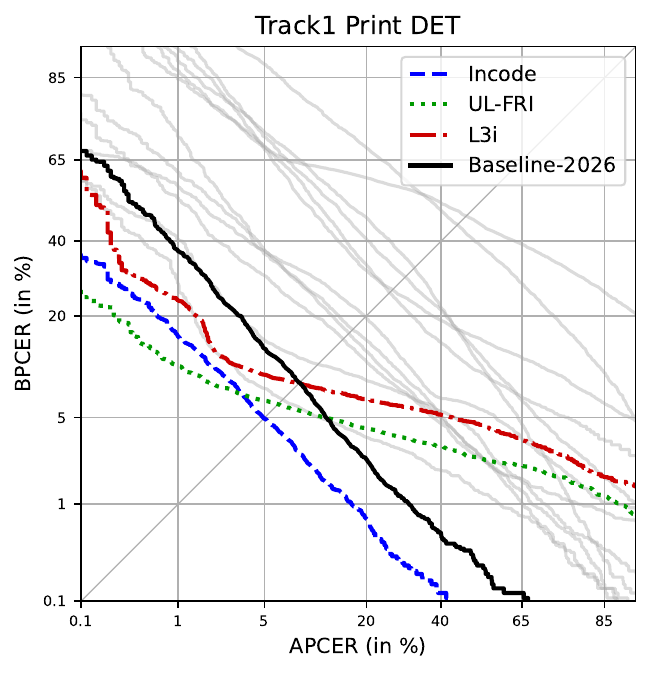}\label{fig:track1-print-det}
}
\subfloat[\textbf{Composite} DET - ID Card + Passport]{
  \includegraphics[width=40mm,height=40mm]{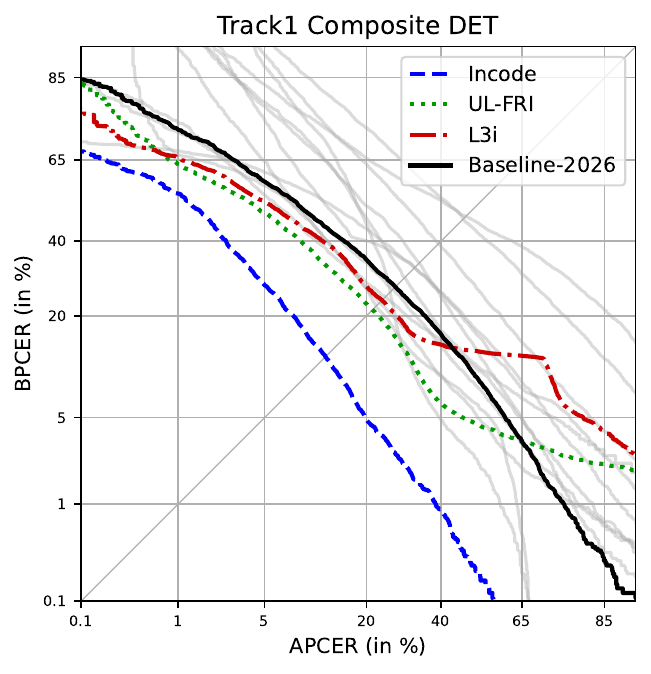}\label{fig:track1-composite-det}}\\
  
% --- Row 2 ---

\subfloat[\textbf{Global} DET - ID Card]{
  \includegraphics[width=40mm,height=40mm]{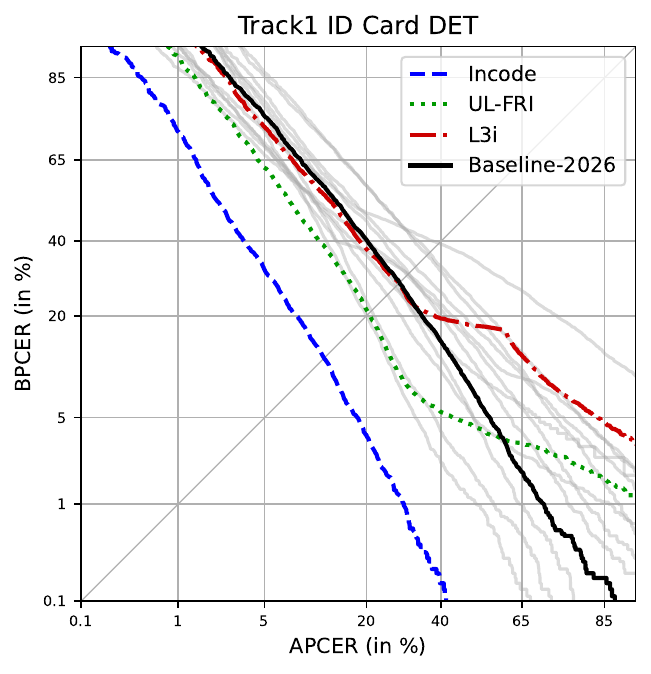}\label{fig:track1-idcard-det}
}
\subfloat[\textbf{Screen} DET - ID Card]{
  \includegraphics[width=40mm,height=40mm]{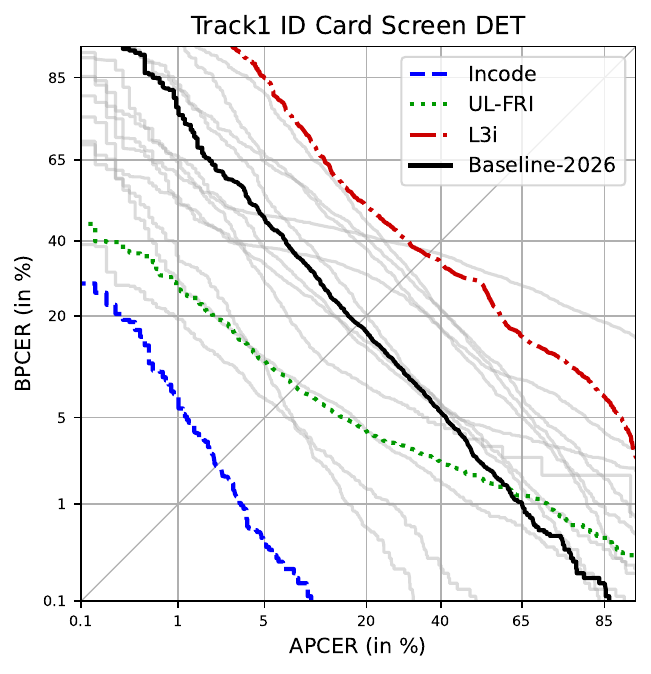}\label{fig:track1-idcard-screen-det}
}
\subfloat[\textbf{Print} DET - ID Card]{
  \includegraphics[width=40mm,height=40mm]{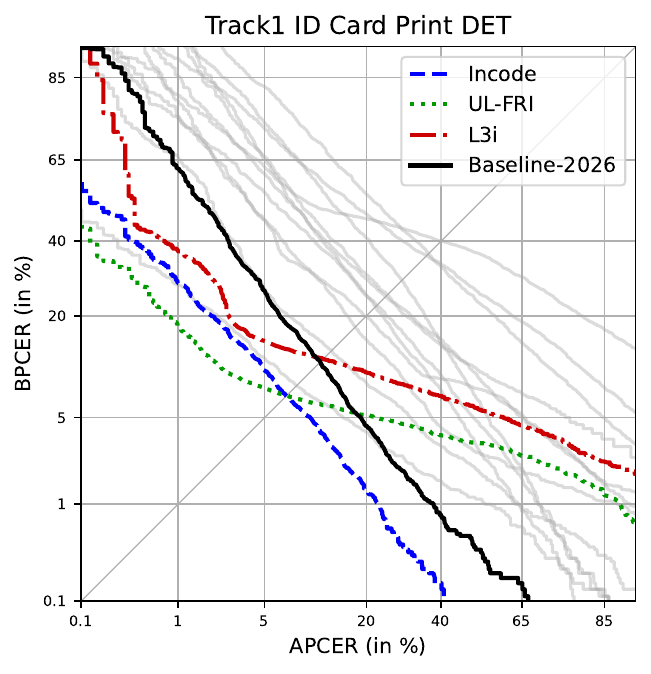}\label{fig:track1-idcard-print-det}
}
\subfloat[\textbf{Composite} DET - ID Card]{
  \includegraphics[width=40mm,height=40mm]{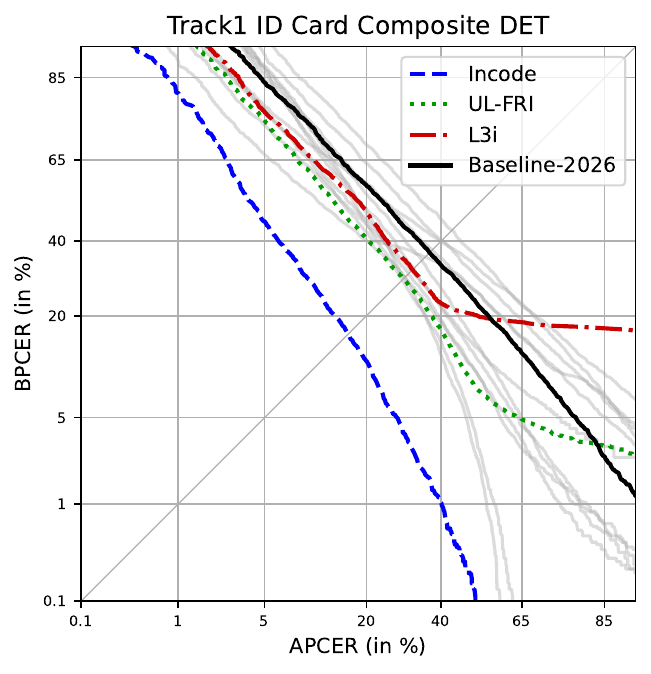}\label{fig:track1-idcard-composite-det}}\\

% --- Row 3 ---

\subfloat[\textbf{Global} DET - Passport]{
  \includegraphics[width=40mm,height=40mm]{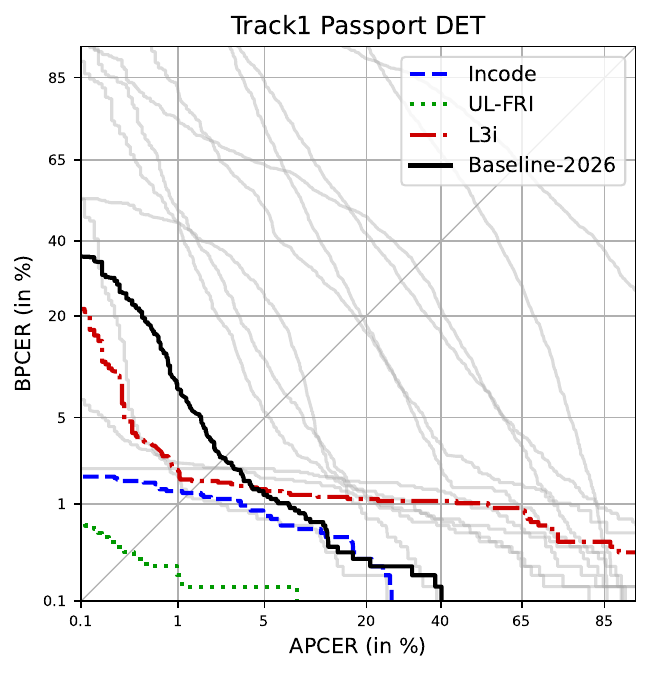}\label{fig:track1-Passport-det}
}
\subfloat[\textbf{Screen} DET - Passport]{
  \includegraphics[width=40mm,height=40mm]{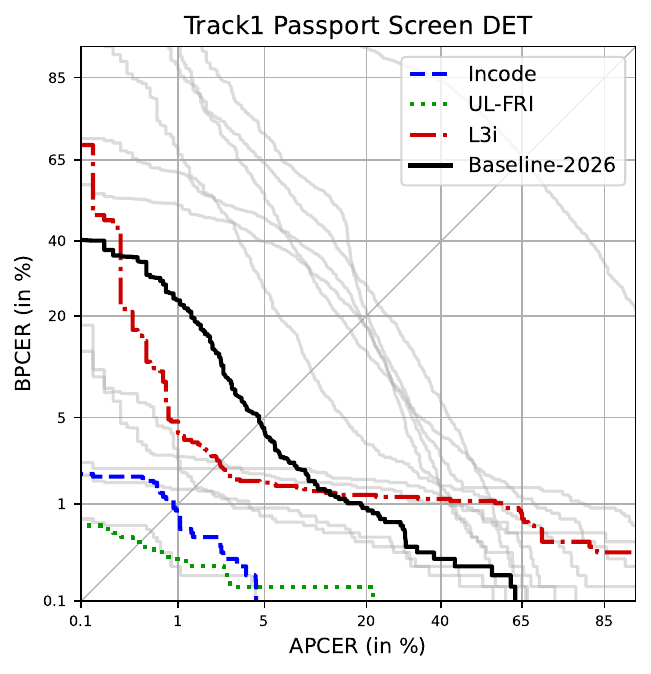}\label{fig:track1-Passport-screen-det}
}
\subfloat[\textbf{Print} DET - Passport]{
  \includegraphics[width=40mm,height=40mm]{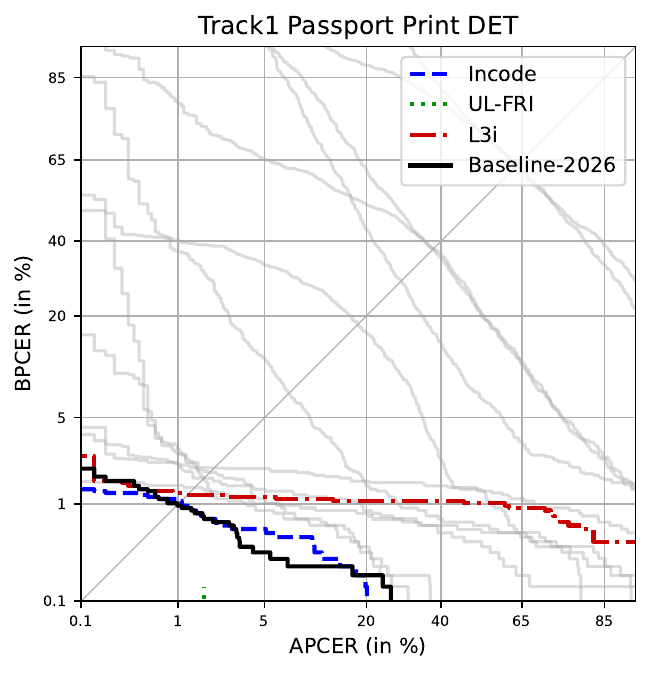}\label{fig:track1-Passport-print-det}
}
\subfloat[\textbf{Composite} DET - Passport]{
  \includegraphics[width=40mm,height=40mm]{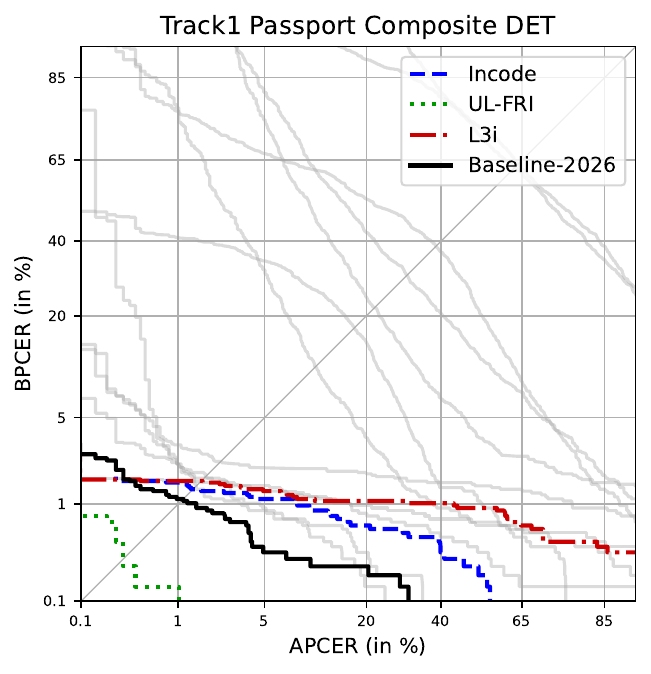}\label{fig:track1-Passport-composite-det}}\\
  
\begin{minipage}[b]{0.24\textwidth} % Empty placeholder to maintain alignment if needed
\end{minipage}
\begin{minipage}[b]{0.24\textwidth}
\end{minipage}

\caption{DET Curves for Track 1. Columns are global, screen, print, and composite attacks. Rows are ID Card plus Passports, only ID Card, and finally only Passports.}
\label{fig:dets-track1}
\end{figure*}

%%%% aqui track 2

% Use figure* for full-page width (top of page)
\begin{figure*}[t]
\centering

% --- Row 1 ---
\subfloat[\textbf{Global} DET - ID Card + Passport]{
  \includegraphics[width=40mm,height=40mm]{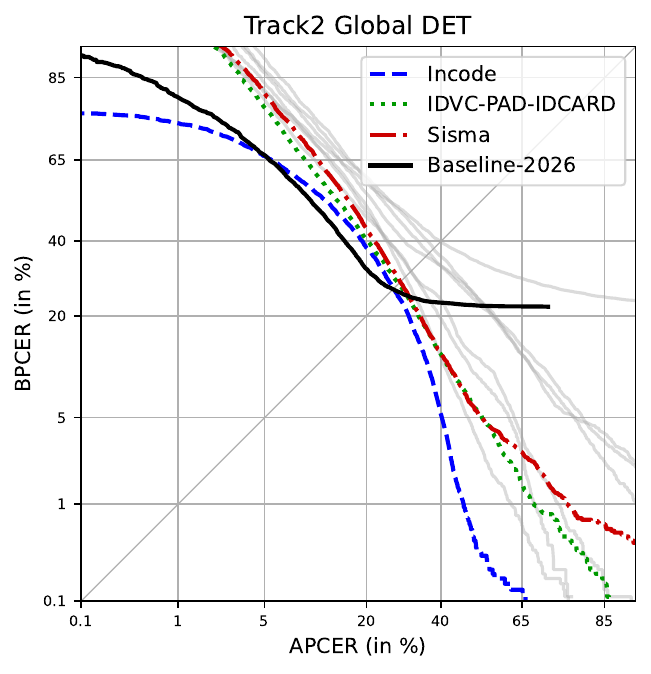}\label{fig:global-det-t2}
}
\subfloat[\textbf{Screen} DET - ID Card + Passport]{
  \includegraphics[width=40mm,height=40mm]{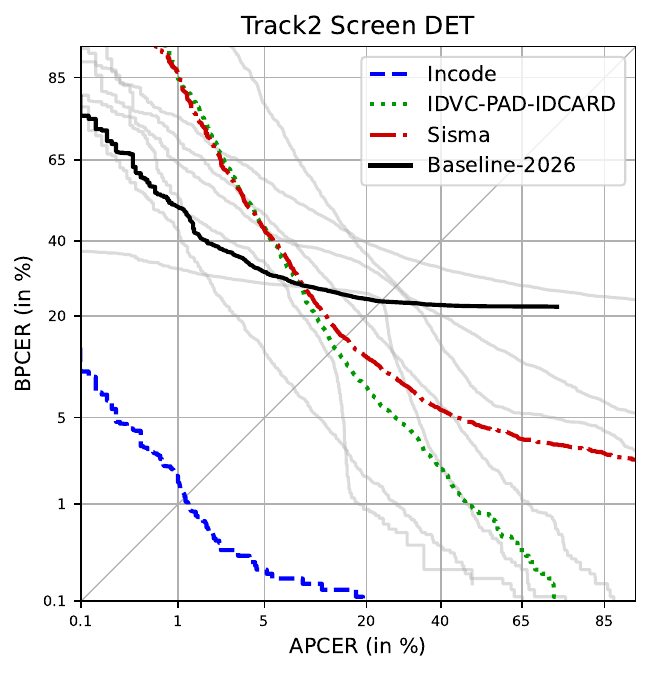}\label{fig:screen-det-t2}
}
\subfloat[\textbf{Print} DET - ID Card + Passport]{
  \includegraphics[width=40mm,height=40mm]{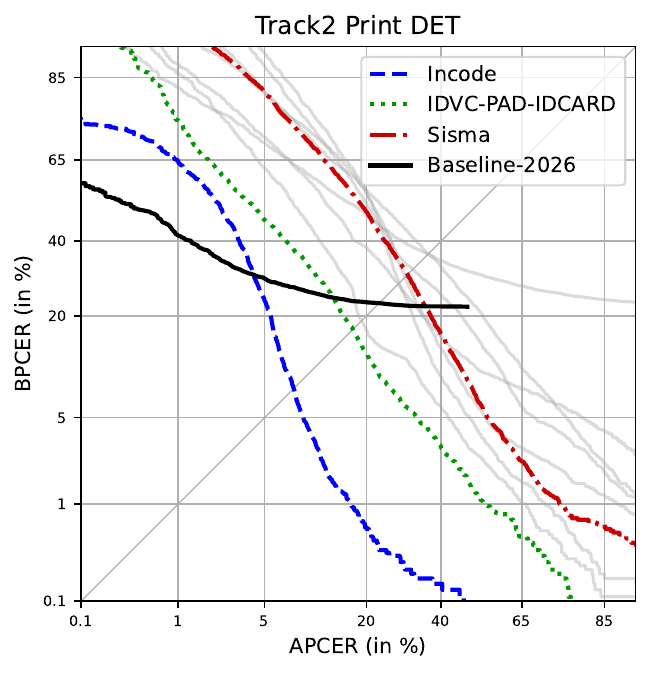}\label{fig:print-det-t2}
}
\subfloat[\textbf{Composite} DET - ID Card + Passport]{
  \includegraphics[width=40mm,height=40mm]{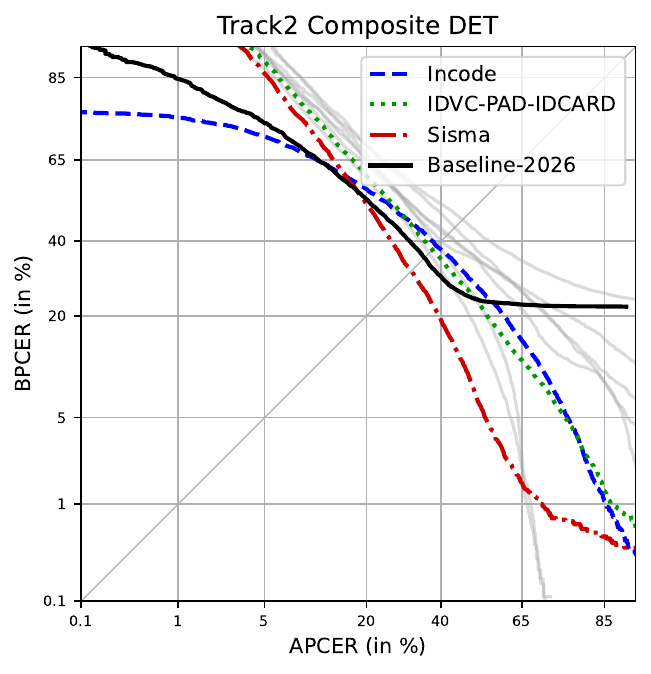}\label{fig:composite-det-t2}}\\
  
% --- Row 2 ---

\subfloat[\textbf{Global} DET - ID Card]{
  \includegraphics[width=40mm,height=40mm]{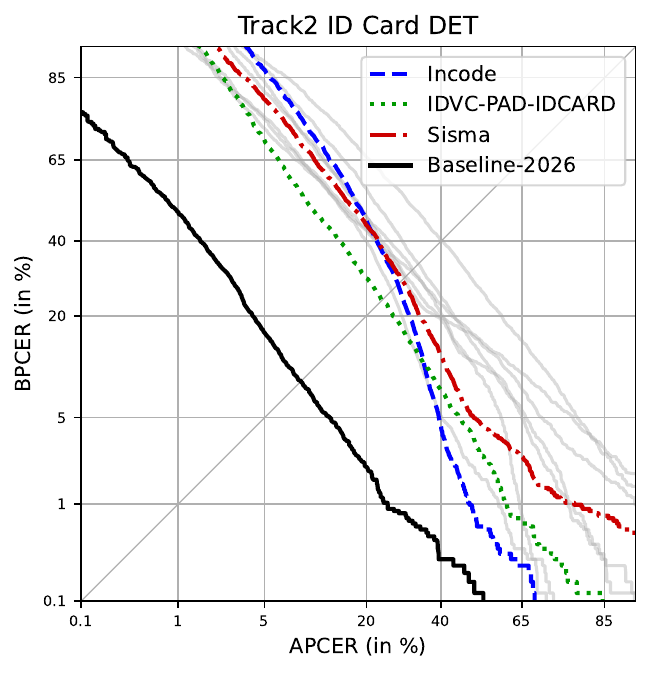}\label{fig:idcard-det-t2}
}
\subfloat[\textbf{Screen} DET - ID Card]{
  \includegraphics[width=40mm,height=40mm]{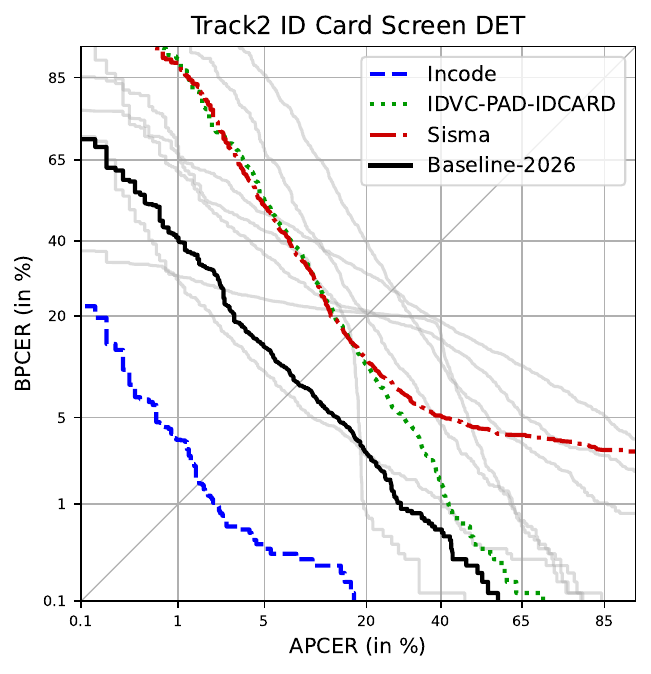}\label{fig:idcard-screen-det-t2}
}
\subfloat[\textbf{Print} DET - ID Card]{
  \includegraphics[width=40mm,height=40mm]{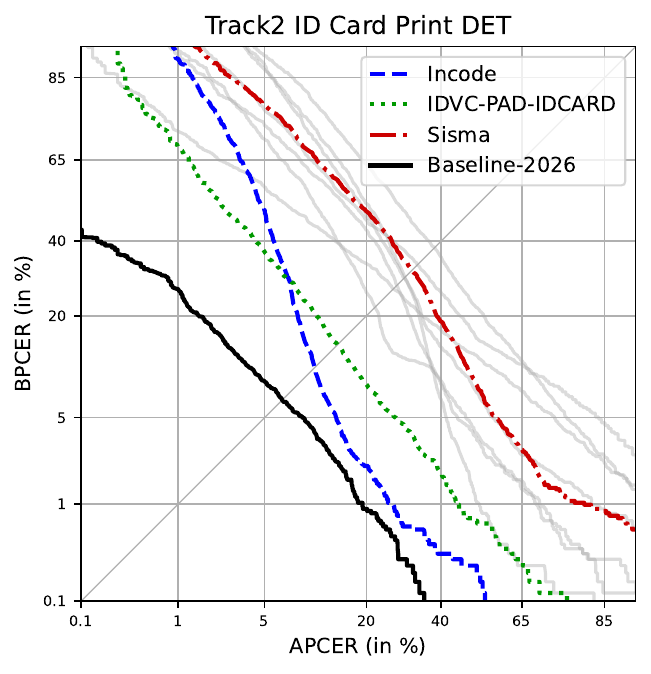}\label{fig:idcard-print-det-t2}
}
\subfloat[\textbf{Composite} DET - ID Card]{
  \includegraphics[width=40mm,height=40mm]{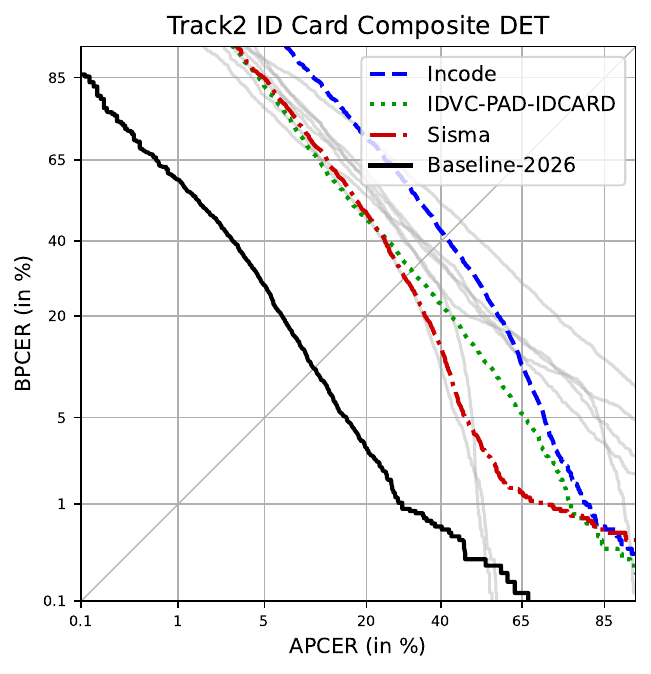}\label{fig:idcard-composite-det-t2}}\\

% --- Row 3 ---

\subfloat[\textbf{Global} DET - Passport]{
  \includegraphics[width=40mm,height=40mm]{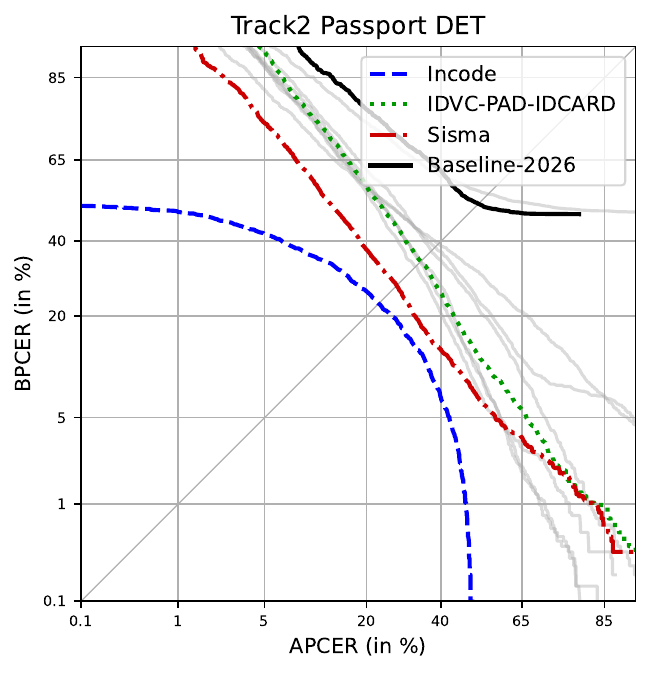}\label{fig:Passport-det-t2}
}
\subfloat[\textbf{Screen} DET - Passport]{
  \includegraphics[width=40mm,height=40mm]{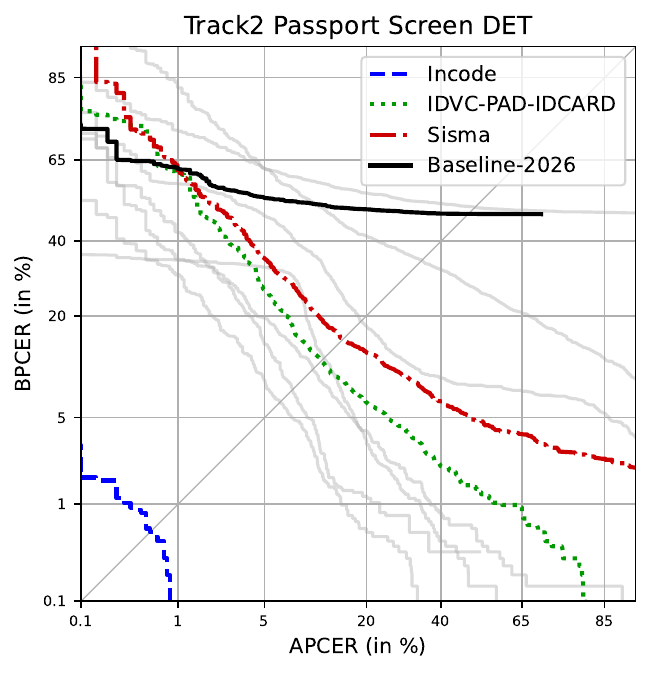}\label{fig:Passport-screen-det-t2}
}
\subfloat[\textbf{Print} DET - Passport]{
  \includegraphics[width=40mm,height=40mm]{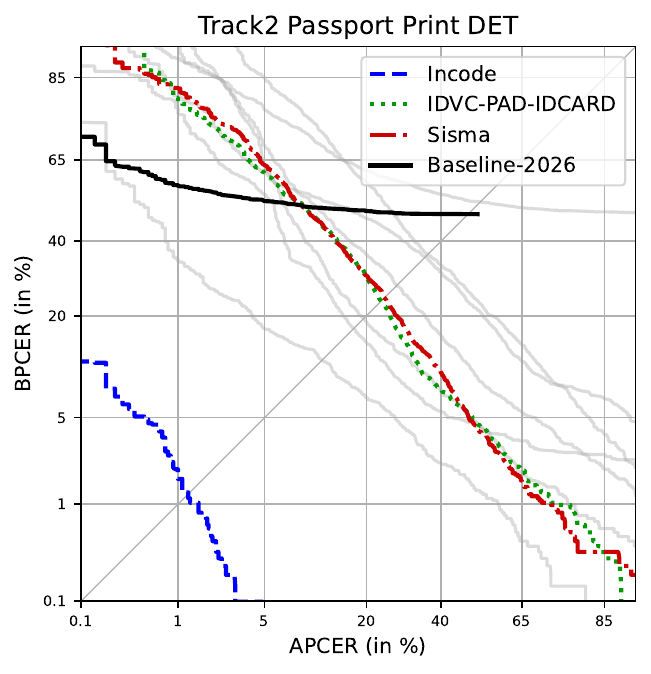}\label{fig:Passport-print-det-t2}
}
\subfloat[\textbf{Composite} DET - Passport]{
  \includegraphics[width=40mm,height=40mm]{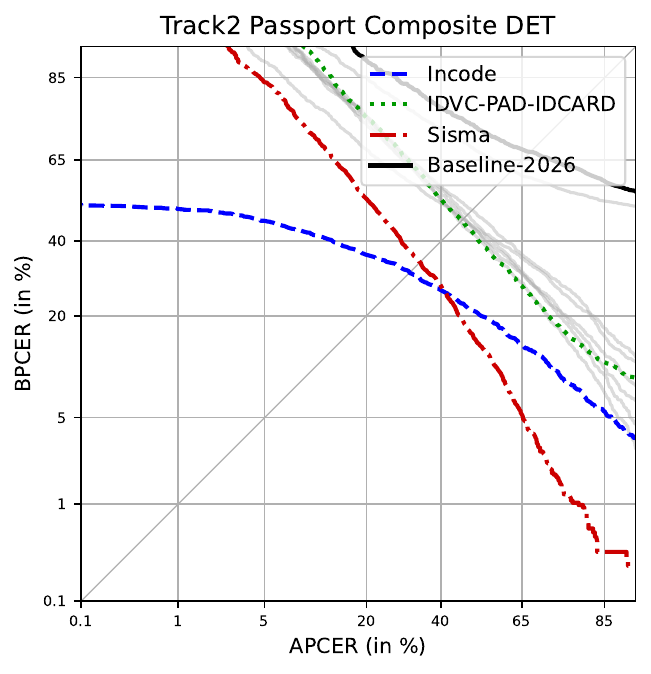}\label{fig:Passport-composite-det-t2}}\\
  
\begin{minipage}[b]{0.24\textwidth} % Empty placeholder to maintain alignment if needed
\end{minipage}
\begin{minipage}[b]{0.24\textwidth}
\end{minipage}

\caption{DET Curves for Track 2. Columns are global, screen, print, and composite attacks. Rows are ID Card plus Passports, only ID Card, and finally only Passports.}
\label{fig:dets-track2}
\end{figure*}

\subsection{Track 1}
The performance analysis for Track 1 highlights the critical challenges of cross-domain generalization when models are trained exclusively on high-fidelity synthetic assets. 

For Track 1, according to our test set, composite attacks are the most difficult attack type. Structurally, ID card samples present a significantly greater level of difficulty than passports.

Figure~\ref{fig:dets-track1} illustrates the DET curves for Track 1. The \textit{Global} DET curve illustrates the best three teams as a binary classifier of bona fide versus all attacks. Figures \ref{fig:track1-screen-det}, \ref{fig:track1-print-det}, and \ref{fig:track1-composite-det} illustrate the results of each team considering bona fide and each attack separately, reported for screen, print, and composite attacks, respectively. Also, Figure~\ref{fig:dets-track1} illustrates, from top to bottom, the DET curves for each type of document, considering ID cards and passports.

\subsection{Track 2}
In this second track, the most complex attack remains a composite attack, which reflects a similar trend to Test 1, in which composites were very difficult to detect. Additionally, overall, detection rates for other types of attacks have worsened, but the error rate by document type has evened out; specifically, the error rates for ID cards and passports are now similar.

Figure~\ref{fig:dets-track2} illustrates the DET curves for Track 2. The \textit{Global} DET curve illustrates the best three teams as a binary classifier of bona fide versus all attacks. Figures \ref{fig:screen-det-t2}, \ref{fig:print-det-t2}, and \ref{fig:composite-det-t2} illustrate the results of each team considering bona fide and each attack separately, reported from screen, print, and composite attacks, respectively. Also, figure~\ref{fig:dets-track2} illustrates, from top to bottom, the DET curves for each type of document, considering ID cards and passports, only ID cards, and only passports.

%%%%%Nuevo parrafo post revision %%%

The results of Track 2 reveal a deeply concerning reality about the current state of PAD systems: even the top-performing team, \textit{Incode}, achieved an average rank ($AVRank$) of $68.71\%$. This result is not competitive but concerning for all the teams participating in \textit{Track 2}. This metric — the average rank across all evaluation criteria (\textit{BPCER10}, \textit{BPCER20}, \textit{BPCER100}) — is lower-is-better; a reliable score would be below $10\%$.
The $68.71\%$ indicates that the top team and other participants exhibit systemic underperformance and a lack of robustness, especially in composite attack.
\textit{\textbf{Incode}}, despite leading in \textit{EER} ($26.52\%$) and several \textit{BPCER} metrics, still ranks below the median in at least one of the four evaluation criteria — likely due to weak performance on \textit{BPCER100 ($75.06\%$)}, which is critical for high-security applications. The fact that no team achieved a lower $AVRank$ underscores a widespread failure to generalize across attack types and thresholds.
Even the \textit{\textbf{Baseline-2026 method}} — a reference framework — scores \textit{\textbf{$AVRank$ = $71.04\%$}}, only slightly worse than \textit{Incode}. This result is not a sign of success but of a fundamental gap in the field: state-of-the-art models do not reliably detect presentation attacks under real-world conditions.
The Baseline-2026 method's near-top-tier performance is not impressive — it is a reminder that even well-designed, open frameworks still fail to deliver strong, consistent results.

\subsection{Conclusion}
\textbf{Track 1} results show \textit{Incode} achieving the best performance with an EER of 8.42\% and BPCER10 of 6.01\%, the lowest in the competition. Its AVRank of 27.82\% confirms superior consistency across all metrics. In contrast, teams like \textit{UL-FRI}, \textit{L3i}, and \textit{ArogyaPandit\_Reasearch\_Team} show increasing vulnerability under high attack pressure ($\text{BPCER100} > 50\%$), while baselines (e.g., Baseline-2026/2025) underperform relative to prior versions, suggesting instability. High AVRanks (e.g., \textit{Pahadis}: 90.80\%) highlight that inconsistent performance leads to poor overall ranking.

\textbf{Track 2} reveals a performance gap between industrial and academic teams, likely due to access to larger proprietary datasets. \textit{Incode} leads with EER = 26.52\%, BPCER10 = 56.53\%, and BPCER100 = 75.06\%, closely followed by \textit{IDVC-PAD-IDCARD} and \textit{Sisma}. Many teams struggle under high attack pressure, exposing a generalization gap.

A key finding is that composite attacks remain the most challenging across both tracks, underscoring the need for robust defense against semantic forgeries. Notably, detection difficulty for ID cards was higher than for passports in Track 1, but this gap diminished in Track 2’s open-set setting.

As future work, we advocate for a deeper investigation into the complexity and resource requirements of Presentation Attack Detection (PAD) models. Many state-of-the-art models from recent competitions are highly complex and computationally demanding, limiting their practicality for real-world deployment—especially in resource-constrained environments such as mobile or edge devices. Finally, our results challenge the assumption that larger, more complex models (e.g., LLM-based) yield better performance. In fact, model size does not correlate with robustness. 

\section*{Disclaimers}
All the models have been evaluated for research purposes only. Any further evaluation needs to be coordinated with the author of these competitions.
The test set is strictly isolated and sequestered to protect all individuals' privacy and identity. As a result, no demographic, gender, or personally identifiable information is included in this report or any public summary. The dataset is not publicly available under any circumstances and is not to be released, redistributed, or reused. All access is handled in accordance with strict confidentiality protocols, and the data is managed in a controlled, secure manner to prevent any risk of re-identification or misuse.

\section*{Acknowledgements} 
This competition was supported by Hochschule Darmstadt (h-da), Facephi, R\&D Department, and Fraunhofer-IGD. Further, this work was supported by the European Union’s Horizon 2020 research and innovation programs under grants 101121280 (EINSTEIN) and CarMen (101168325), and the German Federal Ministry of Education and Research and the Hessian Ministry of Higher Education, Research, Science, and the Arts within their joint support of the National Research Center for Applied Cybersecurity ATHENE.

{\small
\bibliographystyle{ieee}
\bibliography{egbib}
}

\end{document}